\documentclass{article} % For LaTeX2e
\usepackage{iclr2022_conference,times}

\usepackage{amsmath,amsfonts,bm}

\def\eqref#1{equation~\ref{#1}}
\def\1{\bm{1}}

\DeclareMathAlphabet{\mathsfit}{\encodingdefault}{\sfdefault}{m}{sl}
\SetMathAlphabet{\mathsfit}{bold}{\encodingdefault}{\sfdefault}{bx}{n}

\usepackage{hyperref}
\usepackage{url}

\usepackage{xcolor}
\usepackage{amssymb,amsmath}
\usepackage{algorithm}
\usepackage{algorithmicx,algpseudocode}
\usepackage[autostyle]{csquotes}
\usepackage[capitalise]{cleveref}
\usepackage{caption}
\usepackage{subfigure,wrapfig}
\usepackage{amsthm,epsfig,comment}

\newtheorem{theorem}{Theorem}
\newtheorem{assumption}{Assumption}

\newtheorem{lemma}{Lemma}
\newtheorem{proposition}{Proposition}

\usepackage{thmtools,thm-restate}

\expandafter\def\expandafter\normalsize\expandafter{%
    \normalsize
    \setlength\abovedisplayskip{0pt}
    \setlength\belowdisplayskip{0pt}
    \setlength\abovedisplayshortskip{0pt}
    \setlength\belowdisplayshortskip{0pt}
}
\title{model-based offline meta-reinforcement learning with  regularization}

\author{
Sen Lin\textsuperscript{\rm 1},
Jialin Wan\textsuperscript{\rm 1},
Tengyu Xu\textsuperscript{\rm 2},
Yingbin Liang\textsuperscript{\rm 2},
Junshan Zhang\textsuperscript{\rm 1}\\
\textsuperscript{\rm 1} School of ECEE, Arizona State University\\
\textsuperscript{\rm 2} Department of ECE, The Ohio State University\\
\{slin70, jwan20, junshan.zhang\}@asu.edu, \{xu.3260, liang.889\}@osu.edu
}

\iclrfinalcopy % Uncomment for camera-ready version, but NOT for submission.
\begin{document}

\maketitle

\begin{abstract}

Existing offline reinforcement learning (RL) methods face a few major challenges, particularly the distributional shift between the learned policy and the behavior policy. Offline Meta-RL is emerging as a promising approach to address these challenges, aiming to learn an informative meta-policy from a collection of tasks. Nevertheless, as shown in our empirical studies, offline Meta-RL  could be outperformed  by offline single-task RL methods on tasks with good quality of datasets, indicating that a right balance has to be delicately calibrated  between ``exploring" the out-of-distribution state-actions by following the meta-policy and ``exploiting" the offline dataset by staying close to the behavior policy. Motivated by such empirical analysis, we explore model-based offline Meta-RL  with regularized Policy Optimization (MerPO), which learns a meta-model for efficient task structure inference and an informative meta-policy for safe exploration of out-of-distribution state-actions. In particular, we devise a new meta-Regularized model-based Actor-Critic (RAC) method for within-task policy optimization, as
a key building block  of MerPO, using conservative policy evaluation and regularized policy improvement; and the intrinsic tradeoff therein is achieved via striking the right balance between two regularizers, one based on the behavior policy and the other on the meta-policy. We theoretically show that the learnt policy offers guaranteed improvement over both the behavior policy and the meta-policy, thus ensuring the performance improvement on new tasks via offline Meta-RL. Experiments corroborate the superior performance of MerPO over existing offline Meta-RL methods.
\end{abstract}

\section{Introduction}
\vspace{-0.25cm}

Offline reinforcement learning  (a.k.a., batch RL) has recently attracted extensive attention by learning from offline datasets previously collected via some behavior policy \citep{kumar2020conservative}. However, the performance of existing offline RL methods could degrade significantly due to the following issues: 1) the possibly poor quality of offline datasets \citep{levine2020offline} and 2) the inability to generalize to different environments \citep{li2020efficient}. To tackle these challenges, offline {\bf Meta}-RL \citep{li2020multi,dorfman2020offline,mitchell2020offline,li2020efficient} has emerged very recently by leveraging the knowledge of similar offline RL tasks \citep{yu2021conservative}. The main aim of these studies is to enable {\em quick policy adaptation} for new offline tasks, by learning a meta-policy with robust task structure inference that captures the structural properties across training tasks.

\begin{wrapfigure}{r}{.43\textwidth}
%\centering
\vspace{-0.5cm}
\includegraphics[width=0.47\textwidth]{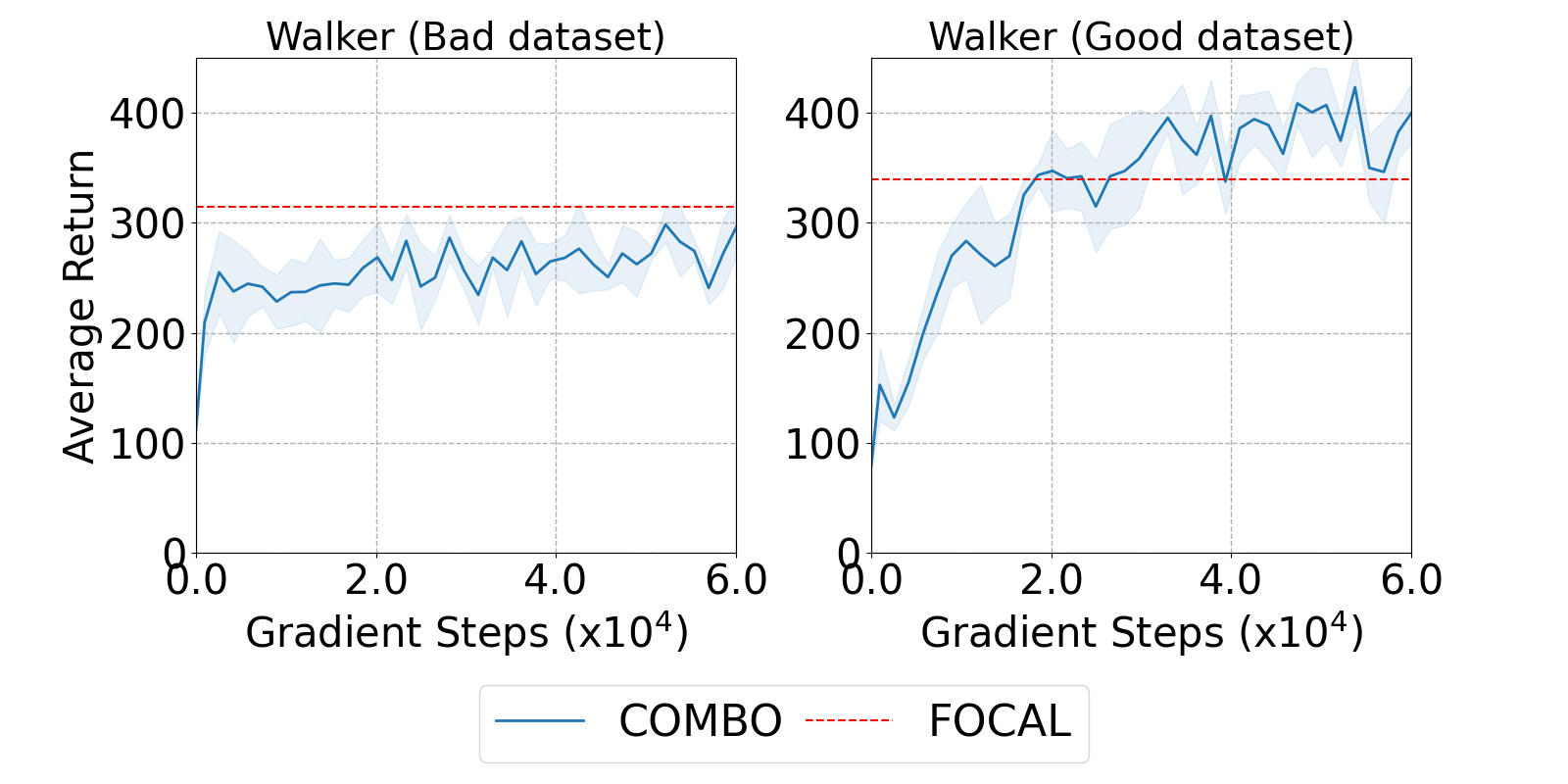}
\vspace{-0.7cm}
\caption{%Performance comparison between 
FOCAL vs.~COMBO.} 
% with behavior-regularized policy improvement 
\label{fig:intro}
\vspace{-0.3cm}
\end{wrapfigure}
Because tasks are trained on offline datasets, value {\em overestimation} \citep{fujimoto2019off} inevitably occurs in dynamic programming based offline Meta-RL, resulted from the distribution shift between the behavior policy and the learnt task-specific policy. To guarantee the learning performance on new offline tasks, a right balance has to be carefully calibrated  between ``exploring" the out-of-distribution state-actions by following the meta-policy, and ``exploiting" the offline dataset by staying close to the behavior policy. 
%With the focus on quick policy adaptation, 
However, such a unique ``exploration-exploitation" tradeoff has not been considered in existing offline Meta-RL approaches, which would likely limit their
ability to handle diverse offline datasets particularly towards those with good behavior policies.
%performance on tasks with a good behavior policy. 
To illustrate this issue more concretely, we compare the performance between a state-of-the-art {\bf offline Meta-RL} algorithm FOCAL~\citep{li2020efficient} and an {\bf offline single-task RL} method COMBO~\citep{yu2021combo} in two new offline tasks. As illustrated in Figure \ref{fig:intro}, while FOCAL performs better than COMBO on the task with a bad-quality dataset (left plot in Figure \ref{fig:intro}), it is outperformed by COMBO on the task with a good-quality  dataset (right plot in Figure \ref{fig:intro}). Clearly, existing offline Meta-RL fails in several standard environments (see \Cref{fig:intro} and \Cref{fig:intro_more}) to generalize universally well over datasets with varied quality.
%In short, existing approaches have not considered this unique ``exploration-exploitation" tradeoff in offline Meta-RL. 
%most of them directly extend the online Meta-RL methods to the offline setup. 
In order to fill such a substantial gap, we seek to answer the following key question in offline Meta-RL:

\emph{How to design an efficient offline Meta-RL algorithm to strike the right balance between exploring with the meta-policy and exploiting the offline dataset?}

To this end, we propose MerPO, a model-based offline {\bf Me}ta-RL approach with {\bf r}egularized {\bf P}olicy {\bf O}ptimization, which learns a meta-model for efficient task structure inference and an informative meta-policy for safe exploration of out-of-distribution state-actions. Compared to existing approaches, MerPO achieves: (1) \emph{safe policy improvement}:   performance improvement can be guaranteed for offline tasks regardless of the quality of the dataset, by strike the right balance between exploring with the meta-policy and exploiting the offline dataset;
%(2) \emph{improved sample efficiency}: sample efficiency is very critical in offline Meta-RL; 
and (2) \emph{better generalization capability}: through a conservative utilization of the learnt model to generate synthetic data, MerPO aligns well with a recently emerging trend in supervised meta-learning to improve the generalization ability by augmenting the tasks with ``more data'' \citep{rajendran2020meta,yao2021improving}.
Our main contributions can be summarized as follows:

(1) Learnt dynamics models not only serve as a natural remedy for task structure inference in offline Meta-RL, but also facilitate better exploration of out-of-distribution state-actions by generating synthetic  rollouts. With this insight, we develop a model-based approach, where an offline meta-model is learnt to enable efficient task model learning for each offline task. More importantly, we propose a meta-regularized model-based actor-critic method (RAC) for within-task policy optimization, 
where a novel regularized policy improvement module is devised to calibrate the unique ``exploration-exploitation" tradeoff  by using an interpolation between two  regularizers, one based on the behavior policy and the other on the meta-policy. Intuitively, RAC generalizes COMBO to the multi-task setting, with introduction of a novel regularized policy improvement module to strike a right balance between the impacts of the meta-policy and the behavior policy.

(2) We theoretically show that under mild conditions, the learnt task-specific policy based on MerPO offers safe performance improvement \emph{over both the behavior policy and the meta-policy} with high probability. Our results also provide a guidance for the algorithm design in terms of how to appropriately select the weights in the interpolation, such that the performance improvement can be guaranteed for new offline RL tasks.

(3) We conduct extensive experiments to evaluate the performance of MerPO. More specifically, the experiments clearly show the safe policy improvement offered in MerPO, corroborating our theoretical results. Further, the superior performance of MerPO over existing offline Meta-RL methods suggests that model-based approaches can be more beneficial in offline Meta-RL.

\section{Related Work}
\vspace{-0.25cm}
\textbf{Offline single-task RL.} 
%As a promising solution to put RL algorithms on the ground by learning effective policies from previously-collected datasets, offline RL has recently garnered much attention.  %To address this issue, a lot of work about both model-free and model-based algorithms have been proposed by following the principle of learning conservatively. 
Many existing model-free offline RL methods  regularize the learnt policy to be close to the behavior policy by, e.g., distributional matching \citep{fujimoto2019off}, support matching \citep{kumar2019stabilizing}, importance sampling \citep{nachum2019algaedice,liu2020off}, learning lower bounds of true Q-values \citep{kumar2020conservative}. Along a different avenue, model-based  algorithms learn policies by leveraging a dynamics model obtained with the offline dataset. \citep{matsushima2020deployment}  directly constrains the learnt policy to the behavior policy as in model-free algorithms.  To penalize the policy for visiting states where the learnt model is likely to be incorrect,  MOPO \citep{yu2020mopo} and MoREL \citep{kidambi2020morel} modify the learnt dynamics such that the value estimates are conservative when the model uncertainty is above a threshold. To remove the need of uncertainty quantification, COMBO \citep{yu2021combo} is proposed by combining model-based policy optimization  \citep{janner2019trust} and conservative policy evaluation \citep{kumar2020conservative}. 

\textbf{Offline Meta-RL.} A few very recent studies have explored the offline Meta-RL. Particularly, \citep{li2020multi} considers a special scenario where the task identity is spuriously inferred due to biased datasets, and applies the triplet loss to robustify the task inference with reward relabelling. \citep{dorfman2020offline} extends an online Meta-RL method VariBAD \citep{zintgraf2019varibad} to the offline setup, and assumes known reward functions and shared dynamics across tasks. Based on MAML \citep{finn2017model}, \citep{mitchell2020offline} proposes an offline Meta-RL algorithm with advantage weighting loss, and learns initializations for both the value function and the policy, where they consider the offline dataset in the format of full trajectories in order to evaluate the advantage. Based on the off-policy Meta-RL method PEARL \citep{rakelly2019efficient}, \citep{li2020efficient} combines the idea of  deterministic context encoder and behavior regularization, under the assumption of deterministic MDP. Different from the above works,  we study a more general offline Meta-RL problem. More importantly,  MerPO strikes a right balance between exploring with the meta-policy and exploiting the offline dataset, which guarantees safe performance improvement for new offline tasks.
% without restricted assumptions. Further,  our approach is based the philosophy of 
%under the theme of improving the performance of new offline tasks by leveraging the knowledge of similar offline tasks. 
% The model-based RL approaches are known to be generally more sample-efficient than the model-free approaches, at the sacrifice of computation efficiency. Nevertheless,
% it is worth to note that sample efficiency is more critical compared to computation efficiency in offline Meta-RL, because all the computation can be simply executed offline in the powerful cloud and guaranteeing that all tasks have enough offline data samples is clearly more challenging. Such a fact substantially increases the role of model-based approaches in offline Meta-RL.

\section{Preliminaries}
\vspace{-0.2cm}

%\subsection{Offline RL and Offline Meta-RL}

Consider a Markov decision process (MDP) $\mathcal{M}=(\mathcal{S}, \mathcal{A}, T, r, \mu_0, \gamma)$ with state space $\mathcal{S}$, action space $\mathcal{A}$, the environment dynamics $T(s'|s,a)$, reward function $r(s,a)$, initial state distribution $\mu_0$, and $\gamma\in (0,1)$ is the discount factor. Without loss of generality, we assume that $|r(s,a)|\leq R_{max}$. Given a policy $\pi$, let $d_{\mathcal{M}}^{\pi}(s):=(1-\gamma)\sum_{t=0}^{\infty}\gamma^t P_{\mathcal{M}}(s_t=s|\pi)$ denote the discounted marginal state distribution, where $P_{\mathcal{M}}(s_t=s|\pi)$ denotes the probability of being in state $s$ at time $t$ by rolling out $\pi$ in $\mathcal{M}$. Accordingly, let $d_{\mathcal{M}}^{\pi}(s,a):=d_{\mathcal{M}}^{\pi}(s)\pi(a|s)$ denote the discounted marginal state-action distribution, and $J(\mathcal{M},\pi):=\frac{1}{1-\gamma}\mathbb{E}_{(s,a)\sim d_{\mathcal{M}}^{\pi}(s,a)}[r(s,a)]$ denote the expected discounted  return. The goal of RL is to find the optimal policy that maximizes $J(\mathcal{M},\pi)$.
In offline RL, no interactions with the environment are allowed, and we only have access to a fixed dataset $\mathcal{D}=\{(s,a,r,s')\}$ collected by some unknown behavior policy $\pi_{\beta}$. Let $d_{\mathcal{M}}^{\pi_{\beta}}(s)$ be the discounted marginal state distribution of $\pi_{\beta}$. The dataset $\mathcal{D}$ is indeed sampled from $d_{\mathcal{M}}^{\pi_{\beta}}(s,a)=d_{\mathcal{M}}^{\pi_{\beta}}(s)\pi_{\beta}(a|s)$. Denote $\overline{\mathcal{M}}$ as the empirical MDP induced by  $\mathcal{D}$ and $d(s,a)$ as a sample-based version of $d_{\mathcal{M}}^{\pi_{\beta}}(s,a)$.

In offline Meta-RL, consider a distribution of RL tasks $p(\mathcal{M})$ as in standard Meta-RL \citep{finn2017model,rakelly2019efficient}, where each task $\mathcal{M}_n$  is an MDP, i.e.,  $\mathcal{M}_n=(\mathcal{S},\mathcal{A},T_n,r_n,\mu_{0,n},\gamma)$, with task-shared state and action spaces, and unknown task-specific dynamics and reward function.% We assume that $\gamma$ is the same for all tasks.
 For each task $\mathcal{M}_n$, no interactions with the environment are allowed and we only have access to an offline dataset $\mathcal{D}_n$, collected by some unknown behavior policy $\pi_{\beta,n}$. The main objective is to learn a meta-policy  based on a set of offline training tasks $\{\mathcal{M}_n\}_{n=1}^N$.
%, which is able to extract the knowledge beyond the offline datasets for improving the performance in new offline RL tasks.

\textbf{Conservative Offline Model-Based Policy Optimization (COMBO).}
Recent model-based offline RL algorithms, e.g., COMBO \citep{yu2021combo}, have demonstrated promising performance on a single offline RL  task
by combining model-based policy optimization \citep{janner2019trust} and conservative policy evaluation (CQL \citep{kumar2020conservative}). 
%To understand how a meta-policy can be leveraged to improve the performance of offline RL tasks, we first outline the basic ideas of COMBO below.
%considering its state-of-the-art performance on existing offline RL benchmarks.
Simply put, COMBO first trains a dynamics model $\widehat{T}_{\theta}(s'|s,a)$ parameterized by $\theta$, via supervised learning on  the offline dataset $\mathcal{D}$. %A model of the reward function can also be learnt in the same manner. 
The learnt MDP is constructed as $\widehat{\mathcal{M}}:=(\mathcal{S},\mathcal{A},\widehat{T},r,\mu_0,\gamma)$.
% A key step in COMBO is policy optimization. 
%Although $\widehat{\mathcal{M}}$ can be directly leveraged to learn a policy via any planning algorithm, model-free methods, e.g., actor-critic \citep{haarnoja2018soft}, are often incorporated to improve the data efficiency, by learning the policy with both $\mathcal{D}$ and model-generated rollouts.
Then, the policy is learnt using $\mathcal{D}$ and model-generated rollouts.
Specifically, define the action-value function (Q-function) as
% {\small
% \begin{align*}
    $Q^{\pi}(s,a):=\mathbb{E}\left[\sum\nolimits_{t=0}^{\infty} \gamma^t r(s_t,a_t)|s_0=s,a_0=a\right]$,
% \end{align*}}%
% and the Bellman operator as:
% %\begin{align*}
%     $\mathcal{B}^{\pi} Q(s,a)=r(s,a)+\gamma \mathbb{E}_{s'\sim T(\cdot|s,a),a'\sim \pi(\cdot|s')}[Q(s',a')]$.
% %\end{align*}
% Since generally it is not possible to obtain the Bellman operator due to limited data, an empirical Bellman operator is computed as a surrogate:
and the empirical Bellman operator as:
%\begin{align*}
    $\widehat{\mathcal{B}}^{\pi} Q(s,a)=r(s,a)+\gamma \mathbb{E}_{(s,a,s')\sim\mathcal{D}}[Q(s',a')]$,
%\end{align*}
for $a'\sim\pi(\cdot|s')$. To penalize the Q functions in out-of-distribution state-action tuples, COMBO employs conservative policy evaluation based on CQL:
% Standard actor-critic methods learn a policy by alternating between two steps: 1) \emph{Policy evaluation}: approximate $Q^{\pi}$ by iterating the Bellman operator:
% %\begin{align*}
%     $\mathcal{B}^{\pi} Q(s,a)=r(s,a)+\gamma \mathbb{E}_{s'\sim T(\cdot|s,a),a'\sim \pi(\cdot|s')}[Q(s',a')]$.
% %\end{align*}
% Since generally it is not possible to obtain the Bellman operator due to limited data, an empirical Bellman operator is computed as a surrogate:
% %\begin{align*}
%     $\widehat{\mathcal{B}}^{\pi} Q(s,a)=r(s,a)+\gamma \mathbb{E}_{(s,a,s')\sim\mathcal{D}}[Q(s',a')]$,
% %\end{align*}
% for $a'\sim\pi(\cdot|s')$.
% 2) \emph{Policy improvement}: improve the policy $\pi(a|s)$ towards actions that maximize the expected Q-value.
% In order to MerPOve the reliance on model uncertainty quantification while still penalizing the value functions for out-of-distribution state-action tuples, COMBO employs a conservative actor-critic algorithm for policy optimization based on CQL:
% (Step 1) \emph{Policy evaluation.} A conservative estimation of $\mathcal{Q}^{\pi}$ is obtained as follows:
{\small
\begin{align}\label{eq:pe_combo}
    \widehat{Q}^{k+1}\leftarrow \arg\min_{Q(s,a)} \beta(\mathbb{E}_{s,a\sim\rho}[Q(s,a)]-\mathbb{E}_{s,a\sim \mathcal{D}}[Q(s,a)])+\frac{1}{2}\mathbb{E}_{s,a,s'\sim d_f}[(Q(s,a)-\widehat{\mathcal{B}}^{\pi}\widehat{Q}^k(s,a))^2]
\end{align}
}%
where $\rho(s,a):=d^{\pi}_{\widehat{\mathcal{M}}}(s)\pi(a|s)$ is the discounted marginal distribution when rolling out $\pi$ in $\widehat{\mathcal{M}}$, and $d_f(s,a)=f d_{\mathcal{M}}^{\pi_{\beta}}(s,a)+(1-f) \rho(s,a)$ for $f\in [0,1]$.  
%Compared to policy evaluation in standard actor-critic methods, an additional term 
%Here $\mathbb{E}_{s,a\sim\rho(s,a)}[Q(s,a)]-\mathbb{E}_{s,a\sim \mathcal{D}}[Q(s,a)]$ is introduced  to push down the Q-values on state-actions in the model rollouts and push up the Q-values on state-actions in $\mathcal{D}$. 
%Following previous model-based methods, e.g., Dyna \citep{sutton1991dyna} and MBPO,
The Bellman backup $\widehat{\mathcal{B}}^{\pi}$ over $d_f$ can be interpreted as an $f$-interpolation of the backup operators under the empirical MDP (denoted by $\mathcal{B}_{\overline{\mathcal{M}}}^{\pi}$) and the learnt MDP (denoted by $\mathcal{B}_{\widehat{\mathcal{M}}}^{\pi}$).
%which captures the trade-off between the sampling error and learnt model bias.
%It can be shown that the learnt Q-function $\widehat{Q}^{\pi}$ by iterating \cref{eq:pe_combo} is a lower bound on the true Q-function $Q^{\pi}$. 
%(Step 2) \emph{Policy improvement.} 
Given the Q-estimation $\widehat{Q}^{\pi}$, the policy can be learnt by:
{\small
\begin{align}\label{eq:pi_combo}
    \pi'\leftarrow \arg\max_{\pi} \mathbb{E}_{s\sim \rho(s), a\sim\pi(\cdot|s)} [\widehat{Q}^{\pi}(s,a)].
\end{align}}%
%Optimizing such a lower bound of $Q^{\pi}$ enables COMBO to safely leverage the learnt model and diverge from the behavior policy, which in turn enhances the potential of achieving higher rewarding policies compared to previous model-free algorithms. 
%\vspace*{-.25in}

\section{MerPO: Model-Based Offline Meta-RL with Regularized Policy Optimization}

%Inspired by the encouraging performance of model-based  offline single-task RL algorithms on solving a single offline RL task, we explore a model-based offline Meta-RL  for improving the performance in new offline RL tasks. Note that the learnt dynamics model serves as a natural remedy for task identification in offline Meta-RL, so as to enable quick policy adaptation.

\begin{figure}
\centering
\includegraphics[scale=0.31]{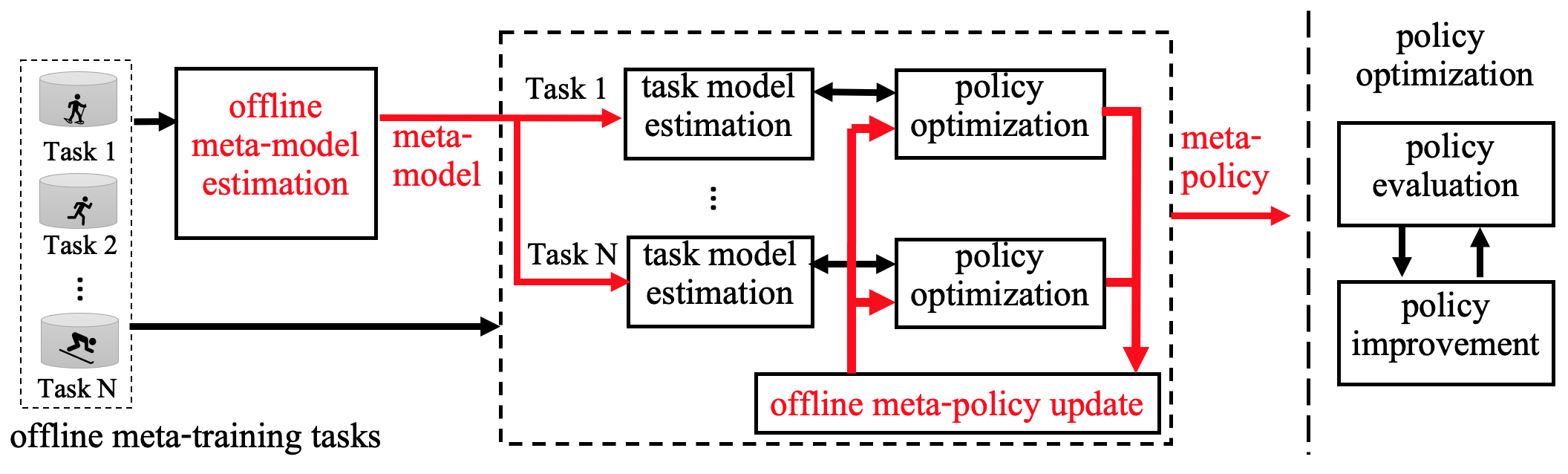}
\caption{Model-based offline Meta-RL with learning of offline meta-model and offline meta-policy.}
\label{fig:metarl}
\end{figure}

Learnt dynamics models not only serves as a natural remedy for task structure inference in offline Meta-RL, but also facilitates better exploration of out-of-distribution state-actions by generating synthetic rollouts \citep{yu2021combo}.
Thus motivated,
%to address the challenges in offline Meta-RL as alluded to earlier,
we propose a general framework of model-based offline Meta-RL, as depicted in Figure \ref{fig:metarl}. More specifically, the offline meta-model is first learnt by using supervised meta-learning, based on which the task-specific model can be quickly adapted. Then, the main attention of this study is devoted to  the learning of an informative meta-policy via bi-level optimization, where 1) a model-based policy optimization approach is leveraged in the inner loop for each task to learn a task-specific policy; and 2) the meta-policy is then updated in the outer loop based on the learnt task-specific policies.

%\vspace*{-.15in}
\subsection{Offline Meta-Model Learning}

Learning a meta-model based on the set of offline dataset $\{\mathcal{D}_n\}_{n=1}^N$ can be carried out via supervised meta-learning. Many gradient-based meta-learning techniques can be applied here, e.g., MAML \citep{finn2017model} and Reptile \citep{nichol2018first}. In what follows, we outline the basic idea to leverage the higher-order information of the meta-objective function. Specifically,  we consider a proximal meta-learning approach, following the same line as in \citep{zhou2019efficient}:
{\small
\begin{align}\label{eq:meta_model}
    \min_{\phi_m}~~L_{model}(\phi_m)=\mathbb{E}_{\mathcal{M}_n}\left\{\min_{\theta_n}~\left[\mathbb{E}_{(s,a,s')\sim \mathcal{D}_n}[\log \widehat{T}_{\theta_n}(s'|s,a)]+\eta \|\theta_n-\phi_m\|^2_2\right]\right\}
\end{align}
}%
where the learnt dynamics for each task $\mathcal{M}_n$ is parameterized by $\theta_n$ and the meta-model is parameterized by $\phi_m$. Solving \cref{eq:meta_model} leads to an offline meta-model.

Given the learnt meta-model $T_{\phi^*_m}$, the dynamics model for an individual offline task $j$ can be found by solving the following problem via gradient descent with initialization  $T_{\phi^*_m}$ using  $\mathcal{D}_j$, i.e.,
{\small
\begin{align}\label{eq:model_local}
    \min_{\theta_j}~~\mathbb{E}_{(s,a,s')\sim \mathcal{D}_j}[\log \widehat{T}_{\theta_j}(s'|s,a)]+\eta \|\theta_j-\phi^*_m\|^2_2.
\end{align}}%
Compared to learning the dynamics model from scratch, adapting from $T_{\phi^*_m}$ can quickly generate a dynamics model for task identity inference by leveraging the  knowledge from similar tasks, and hence improve the sample efficiency \citep{finn2017model,zhou2019efficient}.

\subsection{Offline Meta-Policy Learning}

In this section, we turn attention to tackle one main challenge in this study: How to learn an informative offline meta-policy in order to achieve the optimal tradeoff  between ``exploring" the out-of-distribution state-actions by following the meta-policy and ``exploiting" the offline dataset by staying close to the behavior policy?
%Clearly, value overestimation inevitably occurs in offline Meta-RL as each task is trained on the offline dataset. 
Clearly, it is highly desirable for the meta-policy  to safely `explore' out-of-distribution state-action pairs, and for each task to utilize the meta-policy  to mitigate the issue of value overestimation.

\subsubsection{How Do Existing Proximal Meta-RL Approaches Perform?}

 Proximal Meta-RL approaches have demonstrated remarkable performance in the online setting (e.g., \citep{wang2020global}), by explicitly regularizing the task-specific policy close to the meta-policy. 
 We first consider the approach that applies  the online Proximal Meta-RL method directly to devise offline Meta-RL,
which would lead to:
{\small
\begin{align}\label{eq:standard}
    \max_{\pi_c}~\mathbb{E}_{\mathcal{M}_n}\left\{\max_{\pi_n}\left[\mathbb{E}_{\substack{s\sim \rho_n, \\ a\sim \pi_n(\cdot|s)}} \left[\hat{Q}_n^{\pi}(s,a)\right]-\lambda D(\pi_{n},\pi_c)\right]\right\}
\end{align}}%
where $\pi_c$ is the offline meta-policy, $\pi_n$ is the task-specific policy, $\rho_n$ is the state marginal of $\rho_n(s,a)$ for task $n$ and $D(\cdot,\cdot)$ is some distance measure between two probability distributions. To alleviate value overestimation, conservative policy evaluation can be applied  to learn $\hat{Q}_n^{\pi}$ by using \cref{eq:pe_combo}. Intuitively, \cref{eq:standard} corresponds to generalizing COMBO to the multi-task setting, where a meta policy $\pi_c$ is learned to regularize the within-task policy optimization.

\begin{wrapfigure}{r}{.28\textwidth}
\centering
\vspace{-0.5cm}
\includegraphics[width=0.27\textwidth]{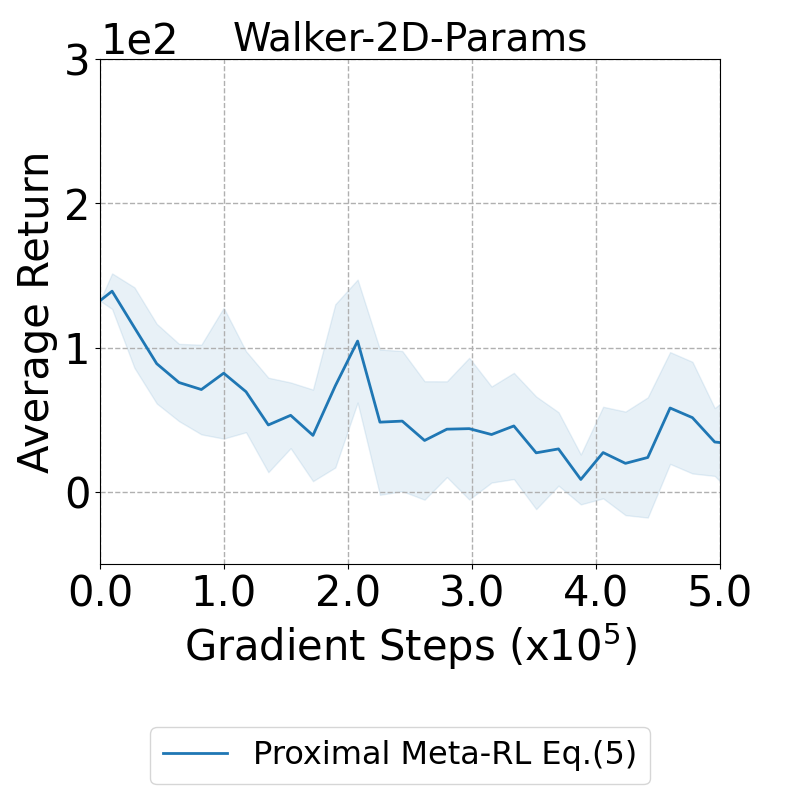}
\vspace{-0.4cm}
\caption{Performance of proximal Meta-RL \cref{eq:standard}.}
\label{fig:std}
\vspace{-0.5cm}
\end{wrapfigure} 
To get a sense of how the meta-policy learnt using \cref{eq:standard} performs, we evaluate its performance in an offline variant of standard Meta-RL benchmark Walker-2D-Params with good-quality datasets, and evaluate the testing performance of the task-specific policy after fine-tuning based on the learnt meta-policy, with respect to the meta-training steps. As can be seen in Figure \ref{fig:std}, the proximal Meta-RL algorithm \cref{eq:standard} performs  surprisingly poorly and  fails to learn an informative meta-policy, despite conservative policy evaluation being applied in within-task policy optimization to deal with the value overestimation. In particular, the testing performance  degrades along with the meta-training process, implying that the quality of the learnt meta-policy is in fact decreasing.

\textbf{\emph{Why does the  proximal Meta-RL method in \cref{eq:standard} perform poorly in offline Meta-RL, even with conservative policy evaluation?}} To answer this, it is worth to take a closer look at the within-task policy optimization in \cref{eq:standard}, which is given as follows:
{\small
\begin{align}\label{eq:pi_test}
    \pi_n\leftarrow \arg\max_{\pi_n} \mathbb{E}_{s\sim \rho_n, a\sim\pi_n(\cdot|s)} [\widehat{Q}_n^{\pi}(s,a)]-\lambda D(\pi_n,\pi_c).
\end{align}}%
Clearly, the performance of  \cref{eq:pi_test} depends heavily  on the quality of the meta-policy $\pi_c$. 
A poor meta-policy may have  negative impact on the performance and result in a task-specific policy $\pi_n$ that is even outperformed by the behaviour policy $\pi_{\beta,n}$. Without online exploration, the quality of $\pi_n$ could not be improved, which in turn leads to a worse meta-policy $\pi_c$ through \cref{eq:standard}. The iterative meta-training process would eventually result in the performance degradation in Figure \ref{fig:std}.

In a nutshell, simply following the meta-policy may lead to worse performance of offline tasks when $\pi_{\beta}$ is a better policy than $\pi_c$. Since it is infeasible to guarantee the superiority of the meta-policy {\em a priori}, it is necessary to balance the tradeoff between exploring with the meta-policy and exploiting the offline dataset, in order to guarantee the performance improvement of new offline tasks.

\subsubsection{Safe Policy Improvement with Meta-Regularization}

To tackle the above challenge, we next devise a novel regularized policy improvement for within-task policy optimization of task $n$, through a weighted interpolation of two different regularizers based on the behavior policy $\pi_{\beta,n}$ and the meta-policy $\pi_c$, given as follows:
{\small
\begin{align}\label{eq:pi_our}
    \pi_n\leftarrow \arg\max_{\pi_n} \mathbb{E}_{s\sim \rho_n, a\sim\pi_n(\cdot|s)} [\widehat{Q}_n^{\pi}(s,a)]-\lambda\alpha D(\pi_n,\pi_{\beta,n})-\lambda(1-\alpha) D(\pi_n,\pi_{c}),
\end{align}}%
for some $\alpha\in [0,1]$. Here, $\alpha$ controls the trade-off between staying close to the behavior policy and following the meta-policy to ``explore'' out-of-distribution state-actions. Intuitively, as $\alpha$ is  closer to $0$, the policy improvement is less conservative and tends to improve the task-specific policy $\pi_n$ towards the actions in $\pi_c$ that have highest estimated Q-values. Compared to \cref{eq:pi_test}, the exploration penalty induced by $D(\pi_n,\pi_{\beta,n})$ serves as a safeguard and 
stops $\pi_n$  following $\pi_c$  over-optimistically. 

\begin{wrapfigure}{r}{.41\textwidth}
\centering
\vspace{-0.8cm}
\begin{minipage}{0.41\textwidth}
\begin{algorithm}[H]	
\small
    \caption{RAC}
	\label{alg1}
 	\begin{algorithmic}[1]
 	  \State Train dynamics model $\widehat{T}_{\theta_n}$ using $\mathcal{D}_n$;
		\For{$k= 1, 2, ...$}
		        \State Perform model rollouts starting from states in $\mathcal{D}_n$ and add into $\mathcal{D}_{model,n}$;
		        \State Policy evaluation by recursively solving \cref{eq:pe_combo} using $\mathcal{D}_n\cup \mathcal{D}_{model,n}$;
		        \State Improve policy by solving \cref{eq:pi_our};
		  \EndFor
	\end{algorithmic}
\end{algorithm}
\end{minipage}
\vspace{-0.4cm}
\end{wrapfigure}
\textbf{Safe Policy Improvement Guarantee.}  Based on conservative policy evaluation \cref{eq:pe_combo} and regularized policy improvement \cref{eq:pi_our}, we have the meta-regularized model-based actor-critic method (RAC), as outlined in Algorithm \ref{alg1}.
Note that different distribution distance measures can be used in \cref{eq:pi_our}. In this work, we theoretically show that the policy $\pi_n(a|s)$ learnt by RAC  is a safe improvement over both the behavior policy $\pi_{\beta,n}$ and the meta-policy $\pi_c$ on the underlying MDP $\mathcal{M}_n$,  when using the maximum total-variation distance for $D(\pi_1,\pi_2)$, i.e., $D(\pi_1,\pi_2):=\max_s D_{TV}(\pi_1||\pi_2)$.

For convenience, define
 $\nu_n(\rho,f)=\mathbb{E}_{\rho} \left[(\rho(s,a)-d_n(s,a))/d_{f,n}(s,a)\right]$, and let $\delta\in (0,1/2)$.
 We have the following important result on the safe policy improvement achieved by $\pi_n(a|s)$.

\begin{theorem}\label{thm1:improve}
(a) Let $\epsilon=\frac{\beta[\nu_n(\rho^{\pi_n},f)-\nu_n(\rho^{\pi_{\beta,n}},f)]}{2\lambda(1-\gamma) D(\pi_n,\pi_{\beta,n})}$. If $\nu_n(\rho^{\pi_n},f)-\nu_n(\rho^{\pi_{\beta,n}},f)>0$ and $\alpha\in \left(\max\{\frac{1}{2}-\epsilon, 0\}, \frac{1}{2}\right)$, then $J(\mathcal{M}_n,\pi_n)\geq \max\{J(\mathcal{M}_n, \pi_c)+\xi_1, J(\mathcal{M}_n, \pi_{\beta,n})+\xi_2\}$ holds with probability at least $1-2\delta$, where both $\xi_1$ and $\xi_2$ are positive for large enough $\beta$ and $\lambda$;

(b)  More generally, we have that $J(\mathcal{M}_n,\pi_n)\geq \max\{J(\mathcal{M}_n, \pi_c)+\xi_1, J(\mathcal{M}_n, \pi_{\beta,n})+\xi_2\}$ holds with probability at least $1-2\delta$,  when $\alpha\in (0,1/2)$,  where $\xi_1$ is positive for large enough $\lambda$.
\end{theorem}

\textbf{Remark 1.} 
The expressions of $\xi_1$ and $\xi_2$ are involved and can be found in \cref{eq:xi1} and \cref{eq:xi2} in the appendix. In part (a) of Theorem \ref{thm1:improve}, both $\xi_1$ and $\xi_2$ are positive for large enough $\beta$ and $\lambda$, pointing to guaranteed  improvements over $\pi_c$ and $\pi_{\beta,n}$.
Due to the fact that the dynamics $T_{\widehat{\mathcal{M}_n}}$ learnt via supervised learning is close to the true dynamics $T_{\mathcal{M}_n}$ on the states visited by the behavior policy $\pi_{\beta,n}$, $d^{\pi_{\beta,n}}_{\widehat{\mathcal{M}_n}}(s,a)$ is close to $d^{\pi_{\beta,n}}_{\overline{\mathcal{M}_n}}(s,a)$ and $\rho^{\pi_{\beta,n}}$ is close to $d_n(s,a)$, indicating that the condition $\nu_n(\rho^{\pi_n},f)-\nu_n(\rho^{\pi_{\beta,n}},f)>0$ is expected to hold in practical scenarios \citep{yu2021combo}. 
For more general cases, a slightly weaker result can be obtained in part (b) of Theorem \ref{thm1:improve}, where $\xi_1$ is positive for large enough $\lambda$ and $\xi_2$ can be negative.

\textbf{Remark 2.}
Intuitively, the selection of $\alpha$   balances the impact of $\pi_{\beta,n}$ and $\pi_c$, while delicately  leaning toward the meta-policy $\pi_c$ because $\pi_{\beta,n}$ has played an important role in policy evaluation to find a lower bound of Q-value. As a result, \emph{\cref{eq:pi_our}    maximizes the true Q-value while implicitly regularized by a weighted combination, instead of $\alpha$-interpolation, between $D(\pi_n,\pi_{\beta,n})$ and $D(\pi_n,\pi_c)$, where the weights are carefully balanced through $\alpha$}. 
%  \begin{wrapfigure}{r}{.42\textwidth}
% \centering
% \vspace{-0.3cm}
% \includegraphics[width=0.42\textwidth]{figure/theorem1.png}
% \caption{Interpretation of Theorem \ref{thm1:improve}.}
% \label{fig:theorem}
% \vspace{-0.2cm}
% \end{wrapfigure}
In particular,
 in the tabular setting, the conservative policy evaluation in \cref{eq:pe_combo} corresponds to penalizing the Q estimation \citep{yu2021combo}:
{\small
\begin{align}\label{eq:penalty}
    \widehat{Q}_n^{k+1}(s,a) = \widehat{\mathcal{B}}^{\pi}\widehat{Q}_n^k(s,a)-\frac{\beta [\rho(s,a)-d_n(s,a)]}{d_{f,n}(s,a)}.
\end{align}}%
Clearly, $\epsilon$ increases with the value of the penalty term in \cref{eq:penalty}. As a result, when the policy evaluation \cref{eq:pe_combo} is overly conservative, the lower bound of $\alpha$ will be close to 0, and hence the regularizer based on the meta-policy $\pi_c$ can play a bigger role so as to encourage the ``exploration'' of out-of-distribution state-actions following the guidance of  $\pi_c$. On the other hand, when the policy evaluation \cref{eq:pe_combo} is less conservative, the lower bound of $\alpha$ will be close to $\frac{1}{2}$, and the regularizer based on $\pi_{\beta,n}$ will have more impact, leaning towards ``exploiting'' the offline dataset. 
%The details are summarized in Figure \ref{fig:theorem}.
In fact, the introduction of 1) behavior policy-based regularizer and 2) the interpolation for modeling the interaction between the behavior policy and the meta-policy, is the key to prove Theorem \ref{thm1:improve}.

\textbf{Practical Implementation.~} In practice, we can use the KL divergence to replace the total variation distance between policies, based on Pinsker's Inequality: $\|\pi_1-\pi_2\|\leq \sqrt{2 D_{KL}(\pi_1||\pi_2)}$.
% \begin{align*}
%     \|\pi_1-\pi_2\|\leq \sqrt{2 D_{KL}(\pi_1||\pi_2)}.
% \end{align*}
Moreover, since the behavior policy $\pi_{\beta,n}$ is typically unknown,   we can use the reverse KL-divergence between $\pi_n$ and $\pi_{\beta,n}$ to circumvent the estimation of $\pi_{\beta,n}$, following the same line as in \citep{fakoor2021continuous}:
{\small
\begin{align*}
    D_{KL}(\pi_{\beta,n}||\pi_n)&=\mathbb{E}_{a\sim \pi_{\beta,n}}[\log \pi_{\beta,n}(a|s)]-\mathbb{E}_{a\sim \pi_{\beta,n}}[\log \pi_n(a|s)]\\
    &\propto -\mathbb{E}_{a\sim \pi_{\beta,n}}[\log \pi_n(a|s)]\approx -\mathbb{E}_{(s,a)\sim \mathcal{D}_n} [\log \pi_n(a|s)].
\end{align*}}%

Then,  the task-specific policy can be learnt by solving the following problem:
{\small
\begin{align}\label{eq:pi_prac}
    \max_{\pi_n}~~\mathbb{E}_{s\sim \rho_n, a\sim \pi_n(\cdot|s)} \left[\widehat{Q}_n^{\pi}(s,a)\right]+\lambda\alpha \mathbb{E}_{(s,a)\sim \mathcal{D}_n} [\log \pi_n(a|s)]-\lambda (1-\alpha)D_{KL}(\pi_n||\pi_c).
\end{align}
}%

\subsubsection{Offline Meta-Policy Update}

 Built on RAC,
the offline meta-policy $\pi_c$ is updated by taking the following two steps, in an iterative manner: 1) (\emph{inner loop}) given the meta-policy $\pi_c$,  RAC is run for each training task to obtain the task-specific policy $\pi_n$; 2) (\emph{outer loop}) based on $\{\pi_n\}_n$,   $\pi_c$ is updated by solving:
{\small
\begin{align}\label{eq:meta-policy}
    \max_{\pi_c}~\mathbb{E}_{\mathcal{M}_n}\left\{\mathbb{E}_{\substack{s\sim \rho_n, \\ a\sim \pi_n(\cdot|s)}} \left[\hat{Q}_n^{\pi}(s,a)\right]+\lambda\alpha \mathbb{E}_{(s,a)\sim \mathcal{D}_n} [\log \pi_{n}(a|s)]-\lambda (1-\alpha)D_{KL}(\pi_{n}||\pi_c)\right\}
\end{align}}%
where  both $\rho_n$ and $\hat{Q}_n^{\pi}$ are from the last iteration of the inner loop for each training task. By using RAC in the inner loop for within-task policy optimization, the learnt task-specific policy $\pi_n$ and the meta-policy $\pi_c$ work in concert to regularize the policy search for each other, and improve akin to  `positive feedback'.
Here the regularizer based on the behavior policy serves an important initial force to boost the policy optimization against the ground: RAC in the inner loop aims to improve the task-specific policy over the behavior policy at the outset and the improved task-specific policy consequently regularizes the meta-policy search as in \cref{eq:meta-policy}, leading to a better meta-policy eventually. Noted that  a meta-Q network is learnt using first-order meta-learning to initialize task-specific Q networks.
It is worth noting that different tasks can have different values of $\alpha$ to capture the heterogeneity of dataset qualities across tasks.

In a nutshell, the proposed model-based offline Meta-RL  with regularized Policy Optimization (MerPO) is built on two key steps: 1) learning the offline meta-model via \cref{eq:meta_model} and 2) learning the offline meta-policy via \cref{eq:meta-policy}. 
%MerPO provides a guided end-to-end model-based solution for offline RL tasks, from meta-regularized task model learning to meta-regularized policy optimization with better sample efficiency.
The details are presented in Algorithm \ref{alg2} in the appendix. 

\subsection{MerPO-based Policy Optimization for New Offline RL Task}

Let  $T_{\phi^*_m}$ and $\pi_c^*$ be the offline meta-model  and the offline meta-policy learnt by MerPO. For a new offline RL task, the task model can be quickly adapted based on \cref{eq:model_local}, and
the task-specific policy can be obtained based on $\pi_c^*$  using the within-task policy optimization module RAC. Appealing to Theorem \ref{thm1:improve}, we have the following result on MerPO-based policy learning on a new task.

\begin{proposition}\label{thm2}
Consider a new offline RL task with the true MDP $\mathcal{M}$.   Suppose
$\pi_o$ is the MerPO-based task-specific policy, learnt by running  RAC over the  meta-policy $\pi_c^*$. If $\epsilon=\frac{\beta[\nu(\rho^{\pi_o},f)-\nu(\rho^{\pi_{\beta}},f)]}{2\lambda(1-\gamma) D(\pi_o,\pi_{\beta})}\geq 0$ and $\alpha\in \left(\max\{\frac{1}{2}-\epsilon, 0\}, \frac{1}{2}\right)$,
then $\pi_o$ achieves the safe performance improvement over both $\pi_c^*$ and $\pi_{\beta}$, i.e., $J(\mathcal{M},\pi_o)> \max\{J(\mathcal{M},\pi_c^*), J(\mathcal{M},\pi_{\beta})\}$ holds with probability at least $1-2\delta$, for large enough $\beta$ and $\lambda$.
\end{proposition}
 Proposition \ref{thm2} indicates that MerPO-based policy optimization for learning task-specific policy  guarantees a policy with higher rewards than both the behavior policy and the meta-policy. This is particularly useful in the following two scenarios:
 1) the offline dataset is collected by some poor behavior policy, but the meta-policy is a good policy; and 2) the meta-policy is inferior to a good behavior policy.
%has nontrivial implications in practice for offline RL to obtain a reliable policy with guaranteed improvement over both the behavior policy and the meta-policy,  when it is not feasible to test any policies in the real environment.

\begin{figure}[t]
\centering
    \subfigure{
         \includegraphics[width=0.23\textwidth]{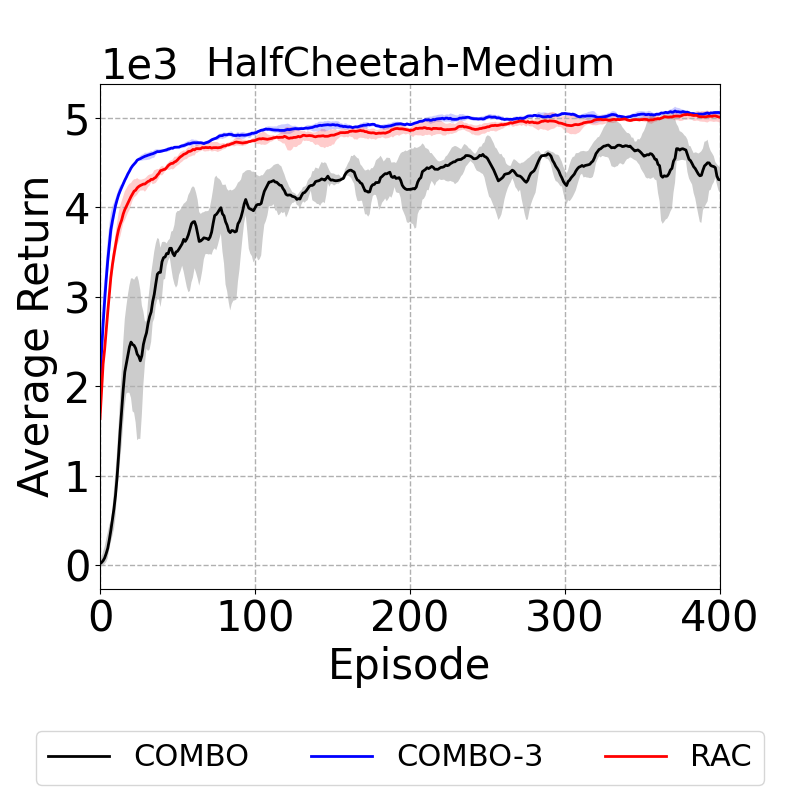}
         \label{fig:hf_good}}
    \subfigure{
         \includegraphics[width=0.23\textwidth]{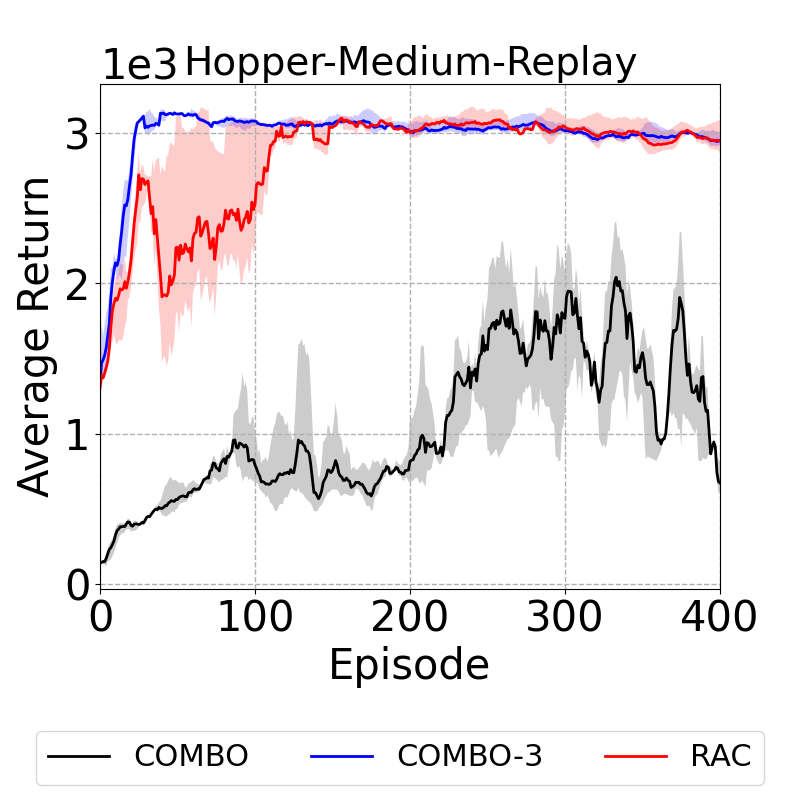}
         %\caption{}
         \label{fig:hop_good}}
    \subfigure{
         \includegraphics[width=0.23\textwidth]{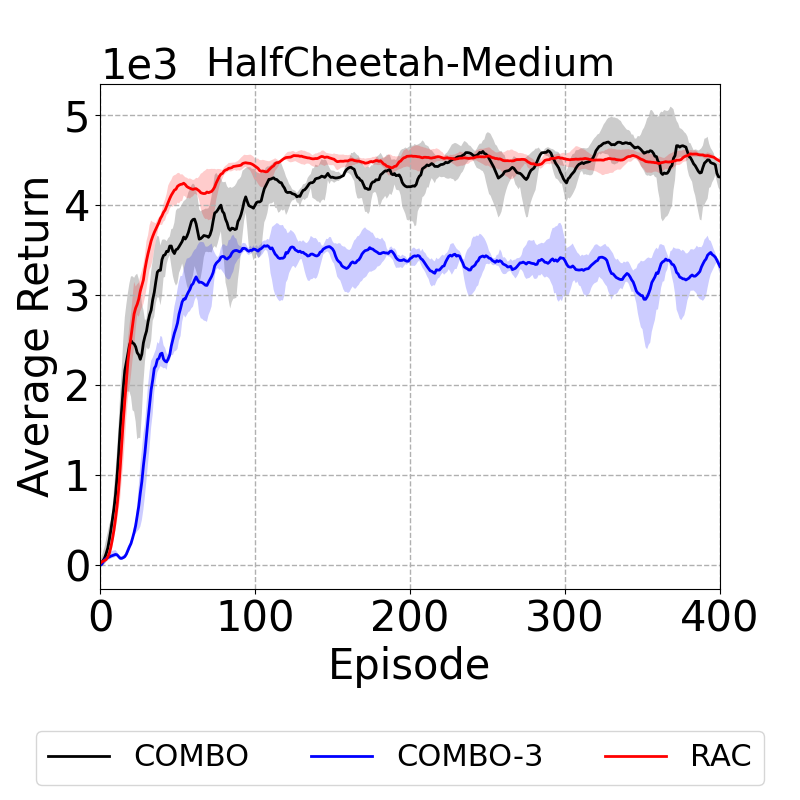}
         %\caption{}
         \label{fig:hf_bad}}
    \subfigure{
         \includegraphics[width=0.23\textwidth]{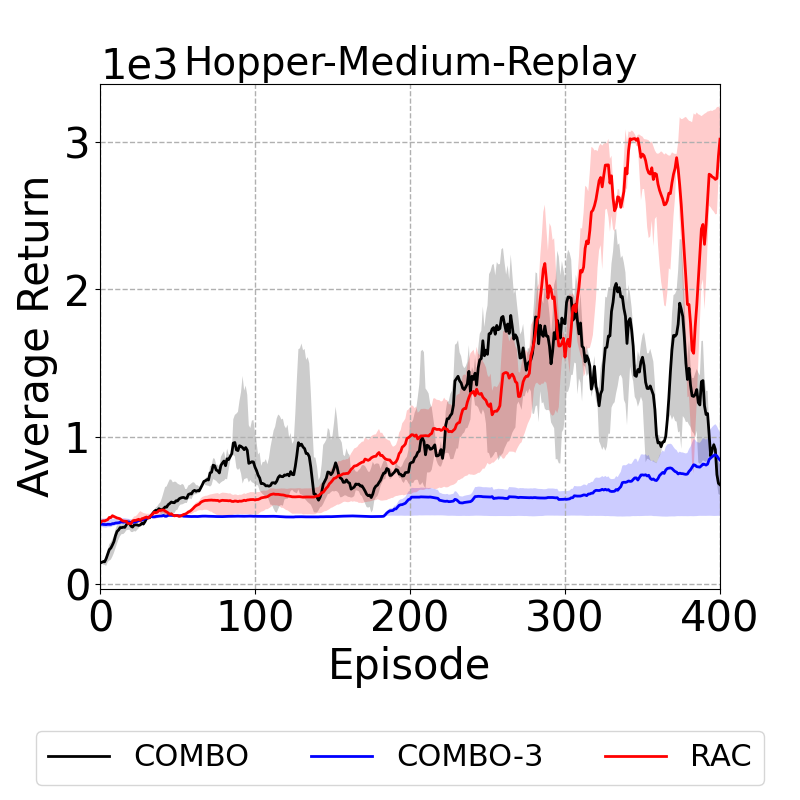}
         %\caption{}
         \label{fig:hop_bad}}       
         \vspace{-0.3cm}
        \caption{Performance comparison among COMBO, COMBO-3 and RAC, with good-quality meta-policy (two figures on the left) and poor-quality meta-policy (two figures on the right). }
         \label{fig:RAC_meta}
\end{figure}

\section{Experiments}

In what follows, we first evaluate the performance of RAC for within-task policy optimization on an offline RL task to validate the safe policy improvement, and then examine how MerPO performs when compared to state-of-art offline Meta-RL algorithms. Due to the space limit, we relegate additional experiments to the appendix.

\subsection{Performance Evaluation of RAC}

\textbf{Setup.~} We evaluate RAC on several continuous control tasks in the D4RL benchmark \citep{fu2020d4rl} from the Open AI Gym \citep{brockman2016openai}, and compare its performance to 1) COMBO (where no meta-policy is leveraged) and 2) COMBO with policy improvement \cref{eq:pi_test} (namely, COMBO-3), under different qualities of offline datasets and different qualities of meta-policy (good and poor). For illustrative purpose, we use a random policy as a poor-quality meta-policy, and choose the learnt policy after 200 episodes as a better-quality meta-policy.
We evaluate the average return over 4 random seeds after each episode with 1000 gradient steps.

\textbf{Results.~} As shown in Figure \ref{fig:RAC_meta}, RAC can achieve comparable performance with COMBO-3 given a good-quality meta-policy, and both clearly outperform COMBO. Besides, the training procedure is also more stable and converges more quickly as expected when regularized with the meta-policy. 
When regularized by a poor-quality meta-policy, that is significantly worse than the behavior policy in all environments, the performance of COMBO-3 degrades dramatically. However,  RAC  outperforms COMBO even when the meta-policy is  a random policy. In a nutshell,  RAC consistently achieves the best performance in various setups and demonstrates compelling robustness against the quality of the meta-policy, for suitable parameter selections ($\alpha=0.4$ in Figure \ref{fig:RAC_meta}).

 \begin{wrapfigure}{r}{.58\textwidth}
          \vspace{-0.5cm}
\centering  
    \subfigure{
\includegraphics[width=0.27\textwidth]{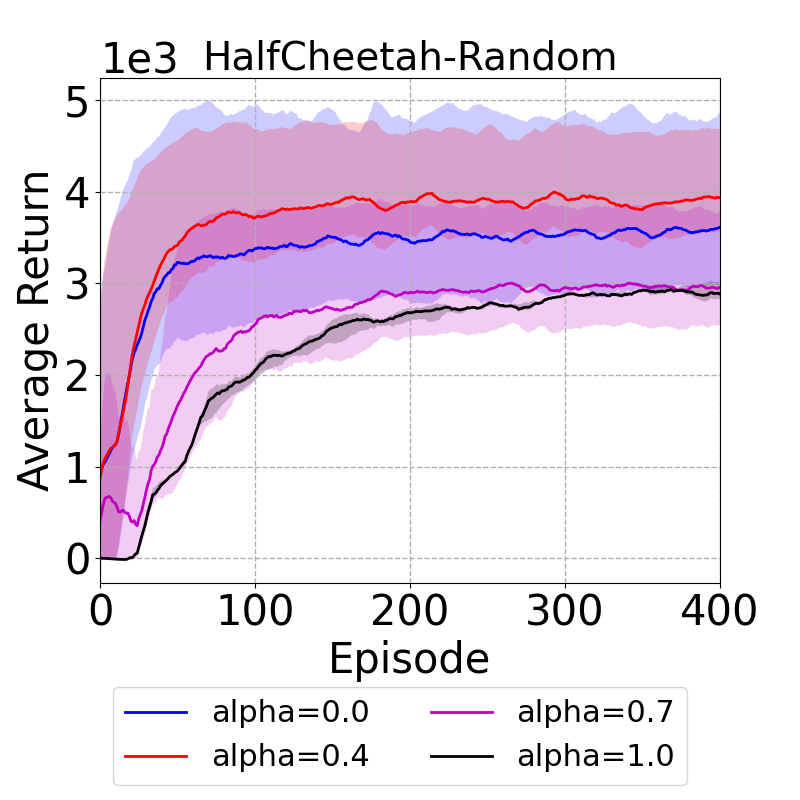}
    \label{fig:hf_rand_alpha}}
     \subfigure{   
\includegraphics[width=0.27\textwidth]{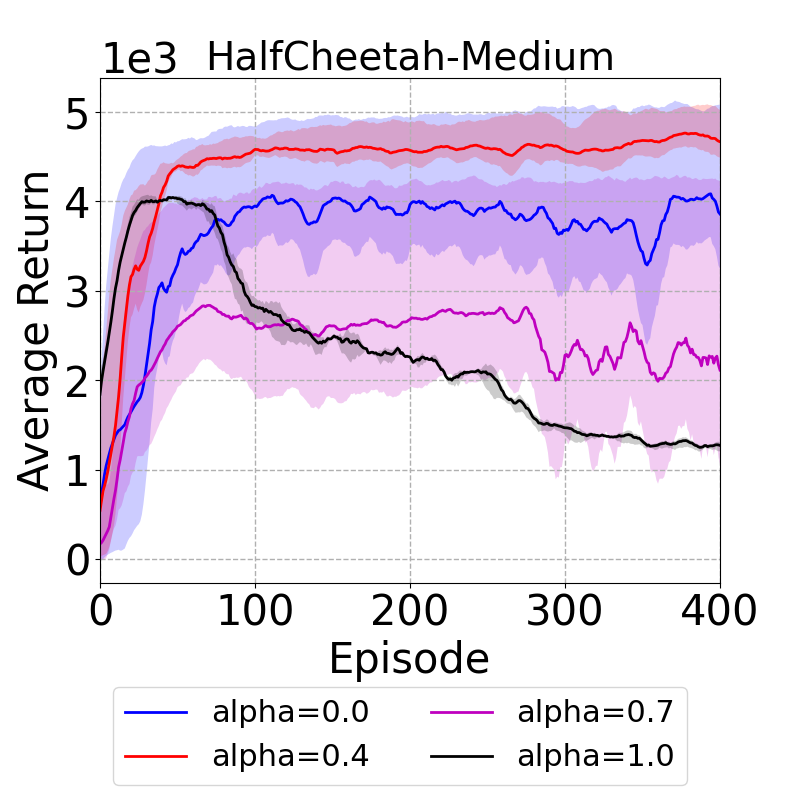}\label{fig:hf_med_alpha}}
 \vspace{-0.5cm}
     \caption{Impact of $\alpha$ on the performance of RAC under different qualities of offline datasets.}
         \label{fig:RAC_alpha_new}
         \vspace{-0.3cm}
\end{wrapfigure}
\textbf{Impact of $\alpha$.~} As shown in Theorem \ref{thm1:improve}, the selection of $\alpha$ is important to guarantee the safe policy improvement property of RAC. Therefore, we next examine the impact of $\alpha$ on the performance of RAC under different qualities of datasets and meta-policy. More specifically, we consider four choices of $\alpha$: $\alpha=0, 0.4, 0.7, 1$. Here, $\alpha=0$ corresponds to COMBO-3, i.e., regularized by the meta-policy only, and the policy improvement step is  regularized by the behavior policy only when $\alpha=1$. Figure \ref{fig:RAC_alpha_new} shows the average return of RAC over different qualities of meta-policies under different qualities of the offline datasets. It is clear that RAC achieves the best performance when $\alpha=0.4$ among the four selections of $\alpha$, corroborating the result in Theorem \ref{thm1:improve}. In general, the performance of RAC is  stable for $\alpha\in [0.3, 0.5]$ in our experiments.
%In other words, RAC clearly demonstrates the robustness against the quality of the meta-policies under different qualities of offine datasets, when $\alpha$ is chosen appropriately as guided by Lemma \ref{thm1:improve}.

\subsection{Performance Evaluation of MerPO}

\textbf{Setup.~} To evaluate the performance of MerPO, we follow the setups in the literature \citep{rakelly2019efficient,li2020efficient} and consider continuous control meta-environments of robotic locomotion. More specifically, tasks has different transition dynamics in Walker-2D-Params and Point-Robot-Wind, and different reward functions in Half-Cheetah-Fwd-Back and Ant-Fwd-Back.
We collect the offline dataset for each task by following  the same line as in \citep{li2020efficient}. 
We consider the following baselines: (1) FOCAL \citep{li2020efficient}, a model-free offline Meta-RL approach based on a deterministic context encoder that achieves the state-of-the-art performance; (2) MBML \citep{li2020multi}, an offline multi-task  RL approach  with metric learning; (3) Batch PEARL, which modifies PEARL \citep{rakelly2019efficient} to train and test from offline datasets without exploration; (4) Contextual BCQ (CBCQ), which is a task-augmented variant of the offline RL algorithm BCQ \citep{fujimoto2019off} by integrating a task latent variable into the state information. We train on a set of offline RL tasks, and evaluate the performance of the learnt meta-policy during the training process on a set of unseen testing offline RL tasks.
%Walker-2D-Params, which is simulated via the MuJoCo simulator \citep{todorov2012mujoco}, has random system parameters and hence requires adaptation across the transition dynamics. Point-Robot-Wind is a variant of Sparse-Point-Robot \citep{li2020efficient}, a 2D navigation problem introduced in \citep{rakelly2019efficient}, where each task is to guide a point robot to navigate  to a specific goal location on the edge of a semi-circle from the origin. In Point-Robot-Wind, each task is affected by a distinct ``wind'' uniformly sampled, and hence differs in the transition dynamics.

\textbf{Fixed $\alpha$ vs  Adaptive $\alpha$}. We consider two implementations of MerPO based on the selection of $\alpha$. 1) MerPO: $\alpha$ is fixed as 0.4 for all tasks; 2) MerPO-Adp: at each iteration $k$, given the task-policy $\pi_n^k$ for task $n$ and the meta-policy $\pi_c^k$ at iteration $k$, we update $\alpha^k_n$ using one-step gradient descent to minimize the following problem.
\begin{align}\label{eq:adp_a}
    \min_{\alpha^k_n}~(1-\alpha^k_n)(D(\pi^k_n,\pi_{\beta,n})-D(\pi^k_n,\pi^k_c)), ~~\text{s.t.}~\alpha^k_n\in [0.1,0.5].
\end{align}
The idea is to adapt $\alpha_n^k$ in order to balance between $D(\pi^k_n,\pi_{\beta,n})$ and $D(\pi^k_n,\pi^k_c)$, because Theorem \ref{thm1:improve} implies that the safe policy improvement can be achieved when the impacts of the meta-policy and the behavior policy are well balanced. Specifically, at iteration $k$ for each task $n$,  $\alpha^k_n$ is increased when the task-policy $\pi_n^k$ is closer to the meta-policy $\pi^k_c$, and is decreased  when $\pi_n^k$ is closer to the behavior policy. Note that $\alpha^k_n$ is constrained in the range $[0.1,0.5]$ as suggested by Theorem \ref{thm1:improve}.

\begin{figure}
\centering
\subfigure{
\includegraphics[width=0.23\textwidth]{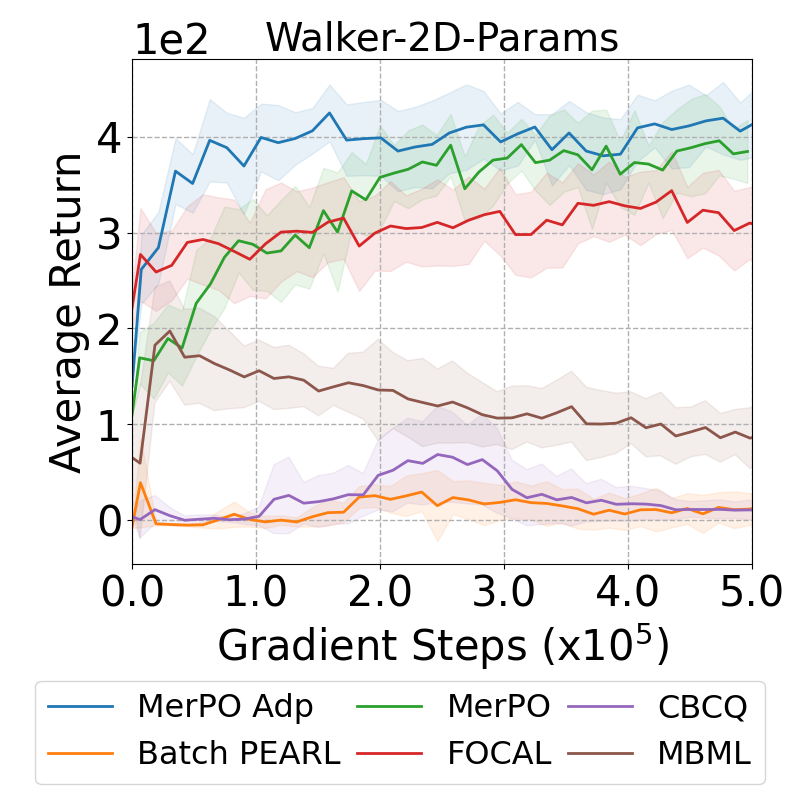}\label{fig:rempo_walker}}
\subfigure{
\includegraphics[width=0.23\textwidth]{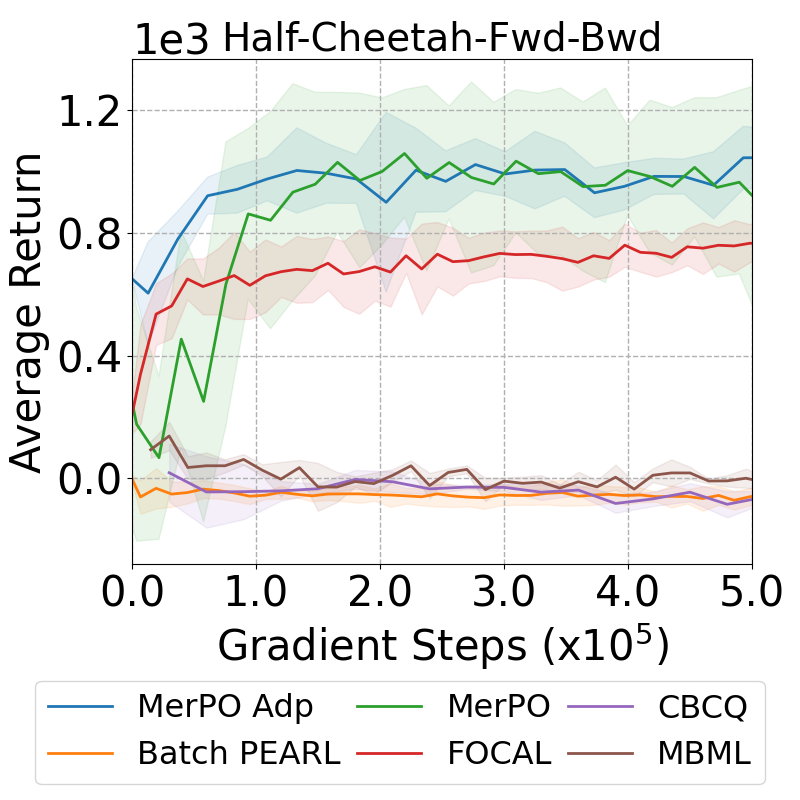}\label{fig:rempo_prw}}
\subfigure{
\includegraphics[width=0.23\textwidth]{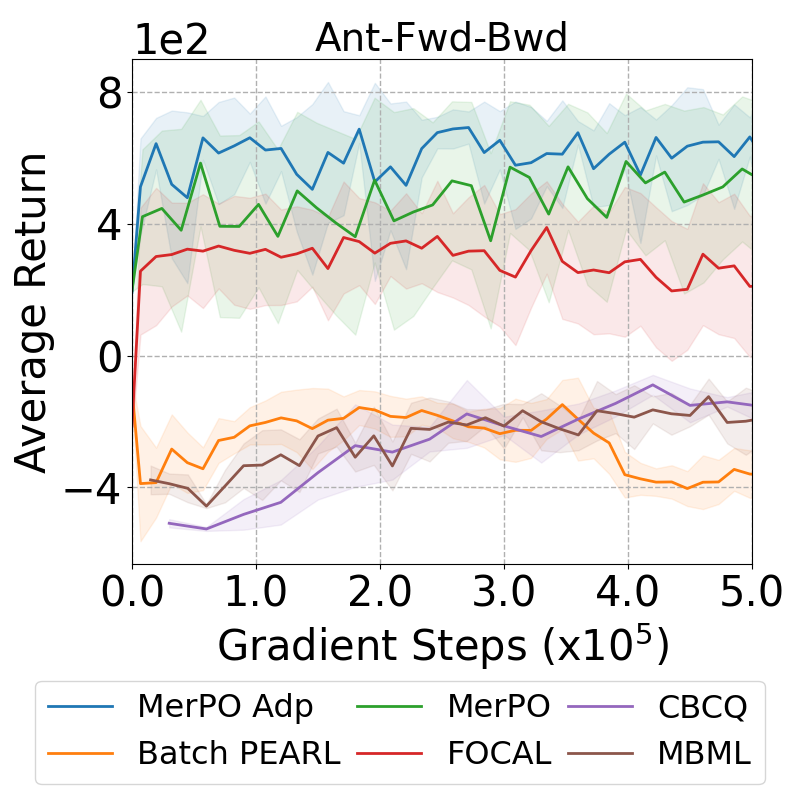}\label{fig:rempo_prw}}
\subfigure{
\includegraphics[width=0.23\textwidth]{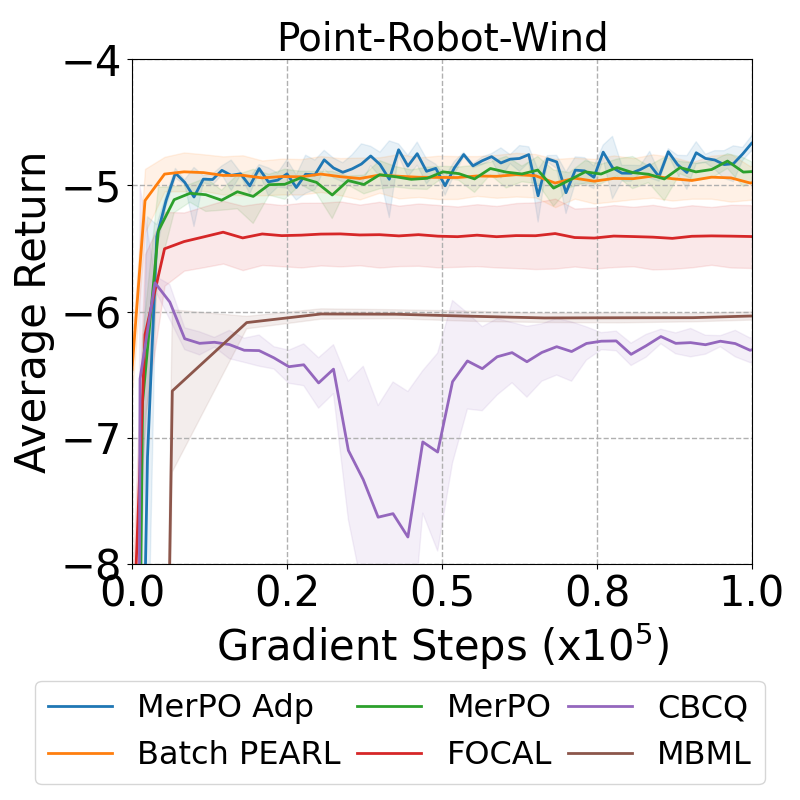}\label{fig:rempo_prw}}
\vspace{-0.3cm}
\caption{Performance comparison in terms of the average return in different environments.}
\vspace{-0.3cm}
\label{fig:rempo}
\end{figure}

\textbf{Results.~} As illustrated in Figure \ref{fig:rempo}, MerPO-Adp yields the best performance, and both MerPO-Adp and MerPO achieve better or comparable performance in contrast to existing offline Meta-RL approaches. Since the meta-policy changes during the learning process and the qualities of the behavior policies vary across different tasks, MerPO-Adp adapts $\alpha$ across different iterations and tasks so as to achieve a `local' balance between the impacts of the meta-policy and the behavior policy. As expected, MerPO-Adp can perform better than MerPO with a fixed $\alpha$. Here the best testing performance for the baseline algorithms is selected over different qualities of offline datasets.

\textbf{Ablation Study.}
We next provide ablation studies by answering the following questions.

\begin{figure}
\centering
\subfigure[Impact of RAC module.]{
\includegraphics[width=0.23\textwidth]{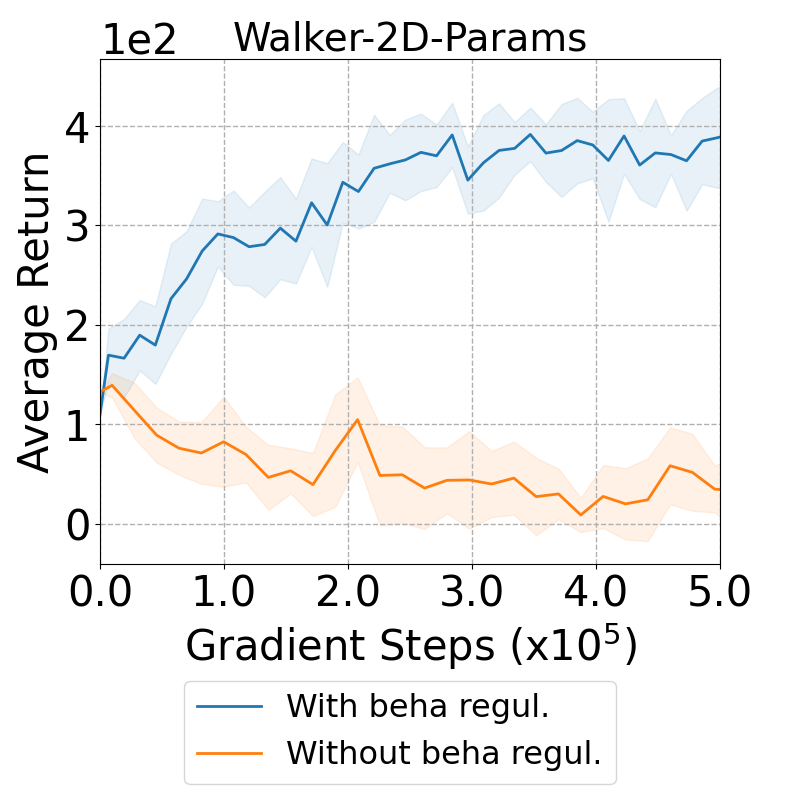}
\label{fig:rempo_beha_new}}
\subfigure[Impact of model utilization.]{
\includegraphics[width=0.23\textwidth]{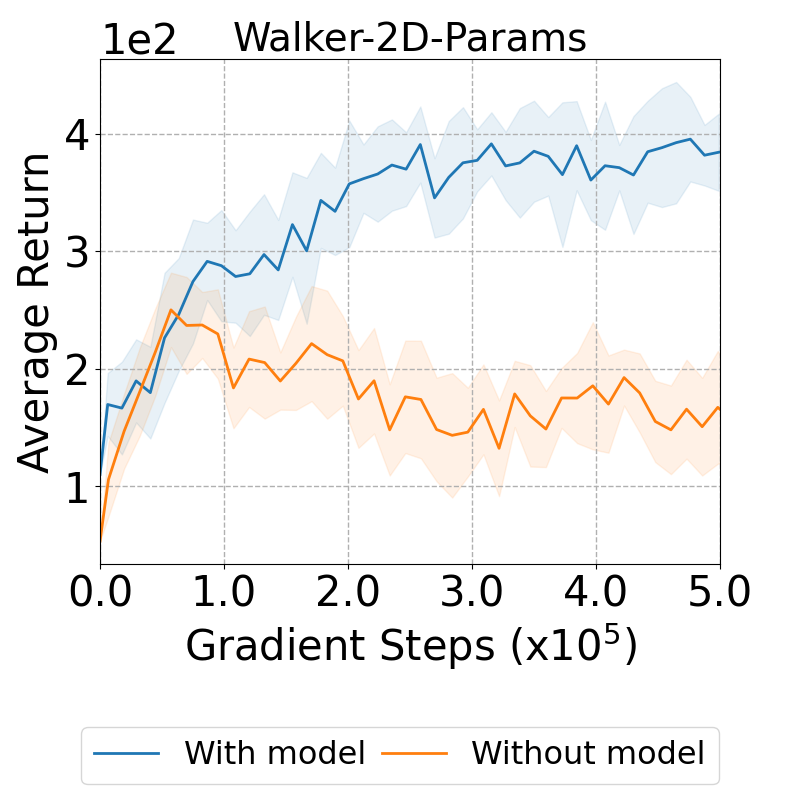}
\label{fig:rempo_model_new}}
\subfigure[Performance under different data qualities.]{
\includegraphics[width=0.23\textwidth]{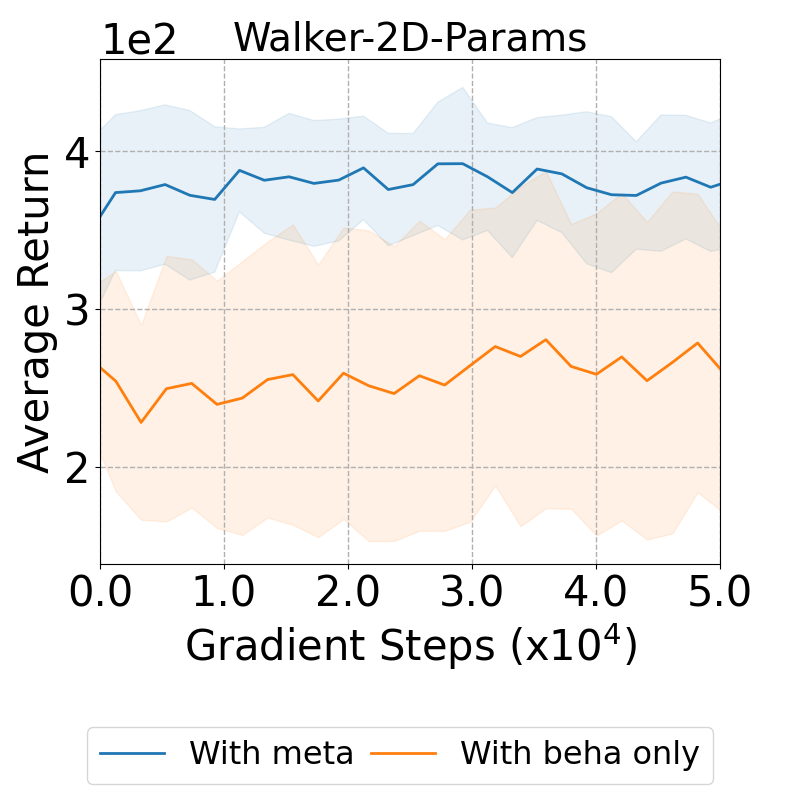}
\label{fig:rempo_test_only_data_new}}
\subfigure[Testing performance for expert dataset.]{
\includegraphics[width=0.23\textwidth]{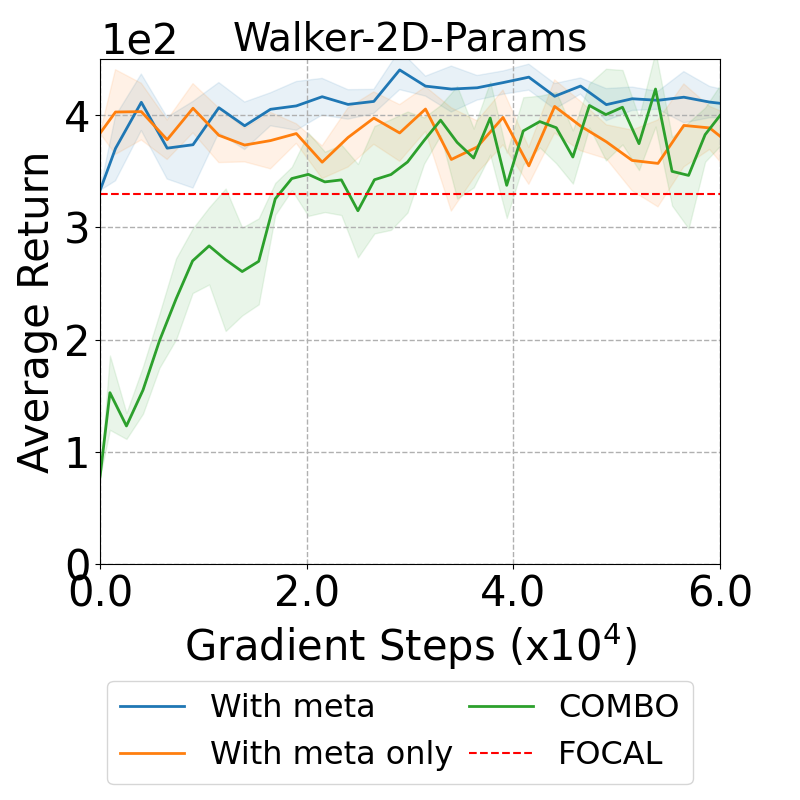}
\label{fig:rempo_walker_exp_new}}
\vspace{-0.3cm}
\caption{Ablation study of MerPO.}
\label{fig:rempo_ablation}
\end{figure}

\textbf{(1) Is RAC important for within-task policy optimization?} 
%Recall that in MerPO each training task exploits RAC to obtain the task-specific policy, where the policy optimization is regularized by both the behavior policy and the meta-policy. 
To answer this question, we compare MerPO with the approach \cref{eq:standard} where the within-task policy optimization is only regularized by the meta-policy. As shown in Figure \ref{fig:rempo_beha_new}, with the regularization based on the behavior policy in RAC, MerPO performs significantly better than \cref{eq:standard}, implying that the safe policy improvement property of RAC enables MerPO to continuously improve the meta-policy.

\textbf{(2) Is learning the dynamics model important?} Without the utilization of models,  the within-task policy optimization degenerates to CQL \citep{kumar2020conservative} and the Meta-RL algorithm becomes a model-free approach. Figure \ref{fig:rempo_model_new} shows the performance comparison between the cases whether the dynamics model is utilized. It can be seen that the performance without model utilization is much worse than that of MerPO. This indeed makes sense because the task identity inference \citep{dorfman2020offline,li2020multi,li2020efficient} is a critical problem in Meta-RL. 
%In MerPO, the task identity can be inferred appropriately based on its dynamics model information. Without model, the tasks can not be well identified, leading to poorlearning performance. 
%This can also be justified by the fact that in model-free offline Meta-RL approaches an additional context encoder is usually exploited to learn the task identity. 
%In fact, by rolling out the policy in the learnt model and generating synthetic data, the generalization ability of the learnt meta-policy can be further improved for unseen offline RL tasks, leading to better performance than the model-free offline Meta-RL methods as illustrated in Figure \ref{fig:rempo}. Such a result 
%is not only consistent with the general consensus that model-based RL approaches make a more natural choice for enabling generalization compared to model-free approaches \citep{yu2020mopo}, but also
Such a result also
aligns well with a recently emerging trend in supervised meta-learning to improve the generalization ability by augmenting the tasks with ``more data'' \citep{rajendran2020meta,yao2021improving}. 

\textbf{(3) How does MerPO perform in unseen offline tasks under different data qualities?} We evaluate the average return in unseen offline tasks with different data qualities, and compare the performance between (1) MerPO with $\alpha=0.4$ (``With meta'') and (2) Run a variant of COMBO with behavior-regularized policy improvement, i.e., $\alpha=1$ (``With beha only''). For a fair comparison, we initialize the policy network with the meta-policy in both cases. As shown in Figure \ref{fig:rempo_test_only_data_new}, the average performance of ``With meta'' over different data qualities is much better than that of ``With beha only''. 
%which showcases the robustness of the proposed algorithm against the quality of offline datasets. 
More importantly, for a new task with expert data, MerPO (``With meta") clearly outperforms COMBO as illustrated in Figure \ref{fig:rempo_walker_exp_new}, whereas the performance of FOCAL is worse than COMBO.

\section{Conclusion}

In this work, we study offline Meta-RL aiming to strike a right balance  between ``exploring" the out-of-distribution state-actions by following the meta-policy and ``exploiting" the offline dataset by staying close to the behavior policy.  To this end, we propose a model-based offline Meta-RL approach, namely MerPO, which learns
a meta-model  to enable efficient task model learning and a meta-policy to facilitate safe exploration of out-of-distribution state-actions. Particularly,
we devise RAC, a meta-regularized model-based actor-critic method  for   within-task policy optimization, by using a weighted interpolation between two regularizers, one based on the behavior policy and the other on the meta-policy. We theoretically show that the learnt task-policy via MerPO offers safe policy improvement over both the behavior policy and the meta-policy.  Compared to existing offline Meta-RL methods, MerPO demonstrates superior performance on several benchmarks, which suggests a more prominent role of model-based approaches in offline Meta-RL.

\newpage

\section*{Acknowledgement}

The work of S. Lin, J. Wan and J. Zhang was supported in part by the U.S. National Science Foundation Grants CNS-2003081, CNS-2203239, CPS-1739344, and CCSS-2121222. The work of T. Xu and Y. Liang was supported in part by the U.S. National Science Foundation Grant DMS-2134145.

\section*{Reproducibility Statement}

For the theoretical results presented in the main text, we state the full set of assumptions of all theoretical results in \cref{ap:pre}, and include the complete proofs of all theoretical results in \cref{ap:proof}. For the experimental results presented in the main text, we include the code in the supplemental material, and specify all the training details in \cref{ap:ex}. For the datasets used in the main text, we also give a clear explanation in  \cref{ap:ex}.

\bibliography{iclr2022_conference}
\bibliographystyle{iclr2022_conference}

\newpage
\appendix

\section{Experimental Details}
\label{ap:ex}

\subsection{Meta Environment Description}
\begin{itemize}
    \item Walker-2D-Params: Train an agent to move forward. Different tasks correspond to different randomized dynamcis parameters.
    \item Half-Cheeta-Fwd-Back: Train a Cheetah robot to move forward or backward, and the reward function depends on the moving direction. All tasks have the same dynamics model but different reward functions.
    \item Ant-Fwd-Back: Train an Ant robot to move forward or backward, and the reward function depends on the moving direction. All tasks have the same dynamics model but different reward functions.
    \item Poing-Robot-Wind: Point-Robot-Wind is a variant of Sparse-Point-Robot \citep{li2020efficient}, a 2D navigation problem introduced in \citep{rakelly2019efficient}, where each task is to guide a point robot to navigate  to a specific goal location on the edge of a semi-circle from the origin. In Point-Robot-Wind, each task is affected by a distinct ``wind'' uniformly sampled from $[-0.05,0.05]^2$, and hence differs in the transition dynamics. 
\end{itemize}

\subsection{Implementation Details and More Experiments}

\subsubsection{Evaluation of RAC}
\label{sec:rac}
\textbf{Model learning.} Following the same line as in \citep{janner2019trust,yu2020mopo,yu2021combo}, the dynamics model for each task is represented as a probabilistic neural network that takes the current state-action as input and outputs a Gaussian distribution over the next state and reward:
\begin{align*}
    \widehat{T}_{\theta}(s_{t+1},r|s,a)=\mathcal{N}(\mu_{\theta}(s_t,a_t),\Sigma_{\theta}(s_t,a_t)).
\end{align*}
An ensemble of 7 models is trained independently using maximum likelihood estimation, and the best 5 models are picked based on the validation prediction error using a held-our set of the offline dataset. Each model is represented by a 4-layer feedforward neural network with 256 hidden units. And one model will be randomly selected from the best 5 models for model rollout.

\textbf{Policy optimization.} We represent both Q-network and policy network as a 4-layer feedforward neural network with 256 hidden units, and use clipped double Q-learning \citep{fujimoto2018addressing} for Q backup update. A max entropy term is also included to the value function for computing the target Q value as in SAC \citep{haarnoja2018soft}.
The hyperparameters used for evaluating the performance of RAC are described in Table \ref{tab:rac}.

\begin{table}
    \caption{Hyperparameters for RAC. }
    \begin{center}
    \begin{tabular}{l|c|c|c}
%         \toprule
\hline
            Hyperparameters     &
          Halfcheetah &
 Hopper & 
           Walker2d \\
          \hline
Discount factor & 0.99 & 0.99 & 0.99\\
Sample batch size & 256 & 256 & 256\\
Real data ratio & 0.5 & 0.5 & 0.5\\
Model rollout length & 5 & 5 & 1 \\
Critic lr & 3e-4 & 3e-4 & 1e-4\\
Actor lr & 1e-4 & 1e-4 & 1e-5\\
Model lr & 1e-3 & 1e-3 & 1e-3\\
Optimizer & Adam & Adam & Adam\\
$\beta$ & 1 & 1 & 10\\
Max entropy & True & True & True \\
$\lambda$ & 1 & 1 & 1\\
            \hline
    \end{tabular}
    \end{center}
    \label{tab:rac}
\end{table}

\begin{figure}[t]
\centering
    \subfigure[Performance in Walker2d with a good meta-policy.]{
         \includegraphics[width=0.23\textwidth]{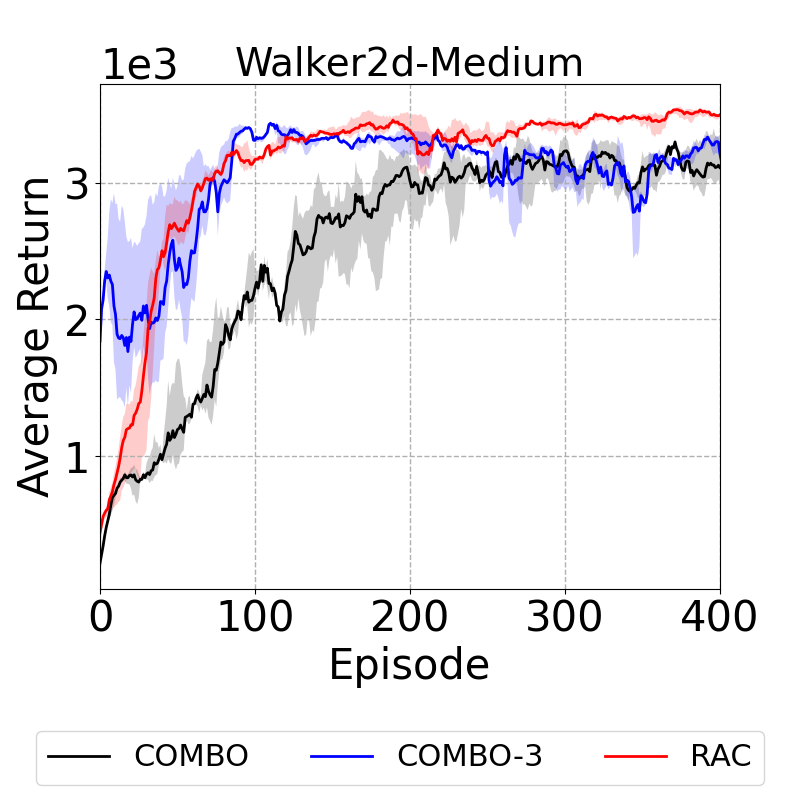}
         %\caption{}
         \label{fig:walker_good}}
    \subfigure[Performance in Walker2d with a random meta-policy.]{
         \includegraphics[width=0.23\textwidth]{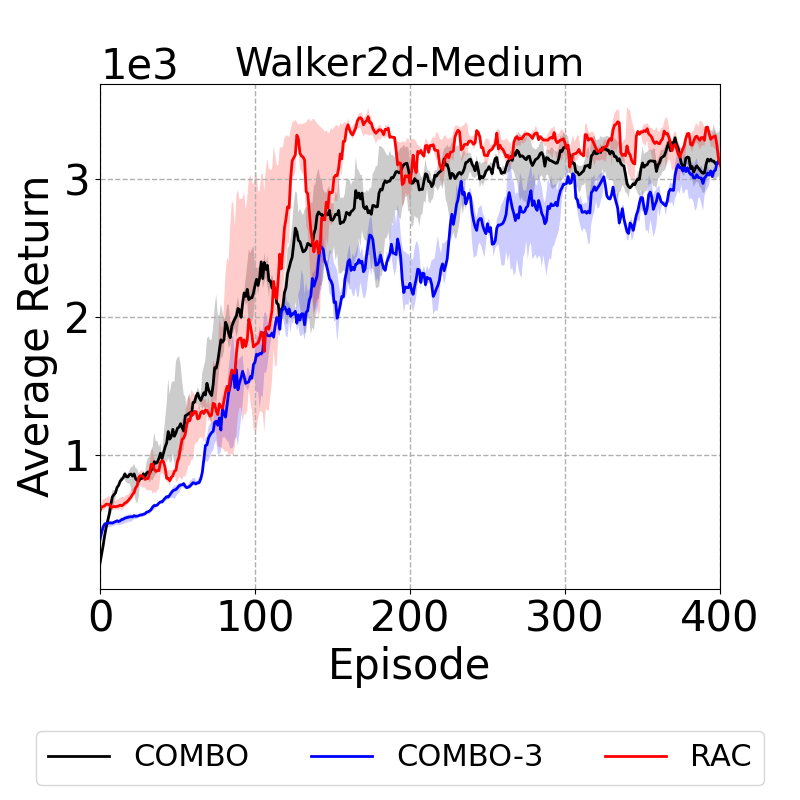}
         %\caption{}
         \label{fig:walker_bad}}
    \subfigure[Performance in HalfCheetah with expert offline dataset.]{
         \includegraphics[width=0.23\textwidth]{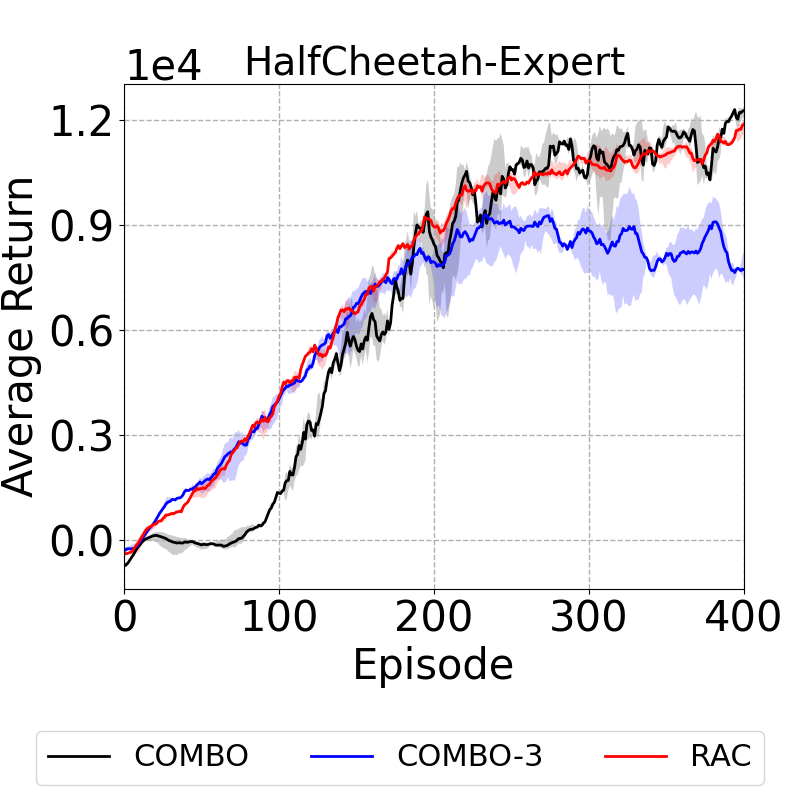}
         %\caption{}
         \label{fig:cheetah_expert}}
            \subfigure[Average return over different qualities of meta-policies under expert dataset for different choices of $\alpha$.]{
         \includegraphics[width=0.23\textwidth]{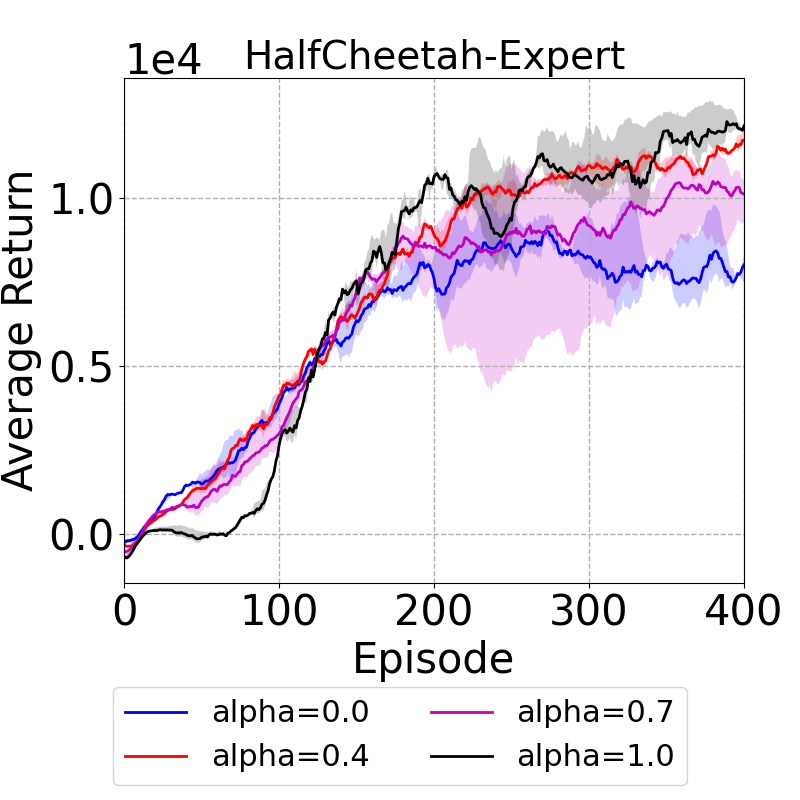}
         %\caption{}
         \label{fig:hf_exp_alpha}}
              \caption{Performance evaluation of RAC. }
         \label{fig:RAC_good}
\end{figure}

\textbf{Additional experiments.} We also evaluate the performance of RAC in Walker2d under different qualities of the meta-policy. As shown in Figure \ref{fig:walker_good} and \ref{fig:walker_bad}, RAC achieves the best performance in both scenarios, compared to COMBO and COMBO-3.
Particularly, the performance of COMBO-3 in Figure \ref{fig:walker_good} degrades in the later stage of training because the meta-policy is not superior over the behavior policy in this case. In stark contrast, the performance of RAC is consistently better, as it provides a safe policy improvement guarantee over both the behavior policy and the meta-policy.

Beside, we also compare the performance of these three algorithms under an expert behavior policy in Figure \ref{fig:cheetah_expert}, where a meta-policy usually interferes the policy optimization and drags the learnt policy away from the expert policy. As expected, RAC can still achieve comparable performance with COMBO, as a result of safe policy improvement over the behavior policy for suitable parameter selections.

We examine the impact of $\alpha$ on the performance of RAC under different qualities of the meta-policy for HalfCheetah with expert data. In this case, the meta-policy is a worse policy compared to the behavior policy. As shown in Figure \ref{fig:hf_exp_alpha}, the performance $\alpha=0.4$ is comparable to the case of $\alpha=1$ where the policy improvement step is only regularized based on the behavior policy, and clearly better than the other two cases.

\subsubsection{Evaluation of MerPO}

\textbf{Data collection.} We collect the offline dataset for each task by training a stochastic policy network using SAC \citep{haarnoja2018soft} for that task and rolling out the policies saved at each checkpoint to collect trajectories. Different checkpoints correspond to different qualities of the offline datasets.
When training with MerPO, we break the trajectories into independent tuples $\{s_i,a_i,r_i,s_i'\}$ and store in a replay buffer. Therefore, the offline dataset for each task does not contain full trajectories over entire episodes, but merely individual transitions collected by some behavior policy.

\textbf{Setup.} For each testing task, we obtain the task-specific policy  through quick adaptation using the within-task policy optimization method RAC, based on its own offline dataset and the learnt meta-policy, and evaluate
the average return of the adapted policy over 4 random seeds. As shown earlier, we take $\alpha=0.4$ for all experiments about MerPO. In MerPO-Adp, we initialize $\alpha$ with 0.4 and update with a learning rate of $1e-4$.

\textbf{Meta-model learning.} Similar as in section \ref{sec:rac}, for each task we quickly adapt from the meta-model to obtain an ensemble of 7 models and pick the best 5 models based on the validation error. The neural network used for representing the dynamics model is same with that in section \ref{sec:rac}.

\textbf{Meta-policy learning.} As in RAC, we represent the task Q-network, the task policy network and the meta-policy network as a 5-layer feedforward neural network with 300 hidden units, and use clipped double Q-learning \citep{fujimoto2018addressing} for within task Q backup update.
For each task, we also use dual gradient descent to automatically tune both the parameter $\beta$ for conservative policy evaluation and the parameter $\lambda$ for regularized policy improvement:
\begin{itemize}
    \item Tune $\beta$. Before optimizing the Q-network in policy evaluation, we first optimize $\beta$ by solving the following problem:
    {\small
\begin{align*}
    \min_{Q}\max_{\beta\geq 0} \beta(\mathbb{E}_{s,a\sim\rho}[Q(s,a)]-\mathbb{E}_{s,a\sim \mathcal{D}}[Q(s,a)]-\tau)+\frac{1}{2}\mathbb{E}_{s,a,s'\sim d_f}[(Q(s,a)-\widehat{\mathcal{B}}^{\pi}\widehat{Q}^k(s,a))^2].
\end{align*}
}%
Intuitively, the value of $\beta$ will be increased to penalty the Q-values for out-of-distribution state-actions if the difference $\mathbb{E}_{s,a\sim\rho}[Q(s,a)]-\mathbb{E}_{s,a\sim \mathcal{D}}[Q(s,a)]$ is larger than some threshold value $\tau$.

    \item Tune $\lambda$. Similarly, we can optimize $\lambda$ in policy improvement by solving the following problem:
        {\small
    \begin{align*}
    \max_{\pi}\min_{\lambda\geq 0} \mathbb{E}_{s\sim \rho(s), a\sim\pi(\cdot|s)} [\widehat{Q}^{\pi}(s,a)]-\lambda\alpha [D(\pi,\pi_{\beta})-D_{target}]-\lambda(1-\alpha) [D(\pi,\pi_{c})-D_{target}].
\end{align*}
}%
Intuitively, the value of $\lambda$ will be increased so as to have  stronger regularizations if the divergence is larger than some threshold $D_{target}$, and decreased if the divergence is smaller than $D_{target}$.
\end{itemize}

Besides, we also build a meta-Q network $Q_{meta}$ over the training process as an initialization of the task Q networks to facilitate the within task policy optimization. At the $k$-th meta-iteration for meta-policy update, the meta-Q network is also updated using the average Q-values of current batch $B$ of training tasks with meta-q learning rate $\xi_q$, i.e., 
\begin{align*}
    Q_{meta}^{k+1} = Q_{meta}^k -\xi_q[Q_{meta}^k-\frac{1}{|B|}\sum_{n\in B} Q_n].
\end{align*}
Therefore, we initialize the task Q networks and the task policy with the meta-Q network and the meta-policy, respectively, for within task policy optimization during both  meta-training and meta-testing. The hyperparameters used in evaluation of MerPO are listed in Table \ref{tab:MerPO}.

For evaluating the performance improvement in a single new offline task, we use a smaller learning rate of $8e-5$ for the Q network and the policy network update.

\begin{table}
\scriptsize
    \caption{Hyperparameters for MerPO. }
    \begin{center}
    \begin{tabular}{l|c|c|c|c}
%         \toprule
\hline
            Hyperparameters     &
          Walker-2D-Params &
          Half-Cheetah-Fwd-Back&
          Ant-Fwd-Back &
           Point-Robot-Wind \\
          \hline
Discount factor & 0.99 & 0.99 & 0.99& 0.9 \\
Sample batch size & 256 & 256 & 256 & 256 \\
Task batch size & 8 & 2 & 2 & 8\\
Real data ratio & 0.5 & 0.5 & 0.5 & 0.5 \\
Model rollout length & 1 & 1 & 1 & 1 \\
Inner critic lr & 1e-3 & 1e-3 & 8e-4 & 1e-3 \\
Inner actor lr & 1e-3 & 5e-4 & 5e-4 & 1e-3 \\
Inner steps & 10 & 10 & 10 & 10\\
Outer critic lr & 1e-3 & 1e-3  & 1e-3 & 1e-3\\
Outer actor lr & 1e-3 & 1e-3& 1e-3 & 1e-3 \\
Meta-q lr & 1e-3 & 1e-3 & 1e-3 & 1e-3\\
Task model lr & 1e-4 & 1e-4 & 1e-4 & 1e-4\\
Meta-model lr & 5e-2& 5e-2& 5e-2& 5e-2 \\
Model adaptation steps & 25& 25& 25& 25\\
Optimizer & Adam & Adam & Adam & Adam\\
Auto-tune $\lambda$ & True & True & True & True\\
$\lambda$ lr & 1 & 1 & 1 & 1 \\
$\lambda$ initial & 5 & 100 & 100 & 5 \\
Target divergence & 0.05 & 0.05 & 0.05 & 0.05 \\
Auto-tune $\beta$ & True & True & True & True \\
$\log\beta$ lr & 1e-3 & 1e-3& 1e-3 & 1e-3 \\
$\log\beta$ initial & 0 & 0 & 0 & 0 \\
Q difference threshold & 5 & 10 & 10 & 10 \\
Max entropy & True & True & True & True\\
$\alpha$ & 0.4 & 0.4 & 0.4 & 0.4\\
Testing adaptation steps & 100 & 100& 100 & 100\\
$\#$ training tasks & 20 &2 &2 & 40\\
$\#$ testing tasks & 5 &2 &2 & 10\\
            \hline
    \end{tabular}
    \end{center}
    \label{tab:MerPO}
\end{table}

\begin{figure}
\centering
\subfigure[Impact of real data ratio.]{
\includegraphics[width=0.31\textwidth]{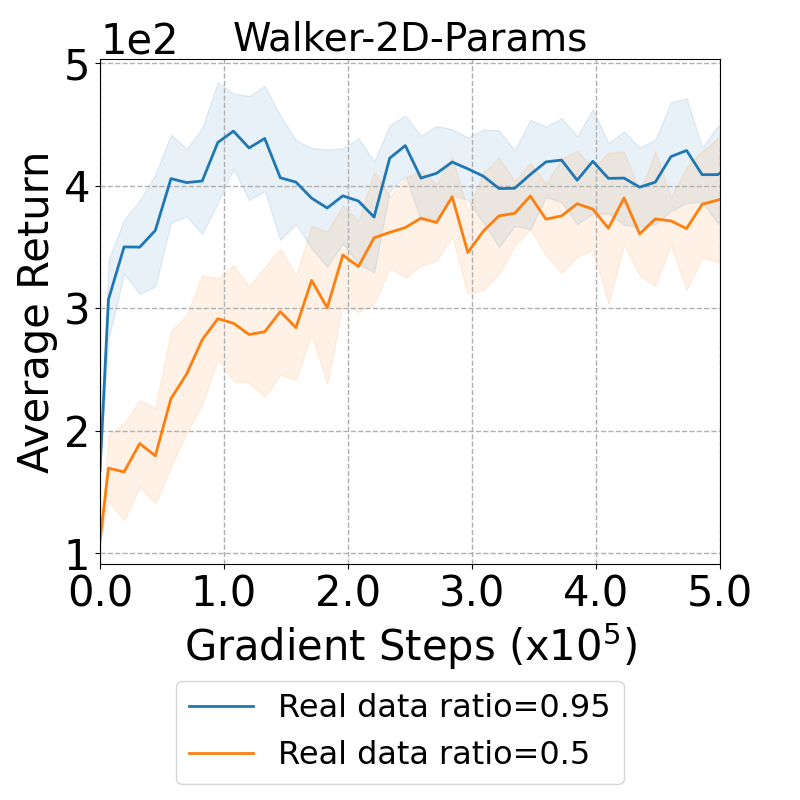}
\label{fig:rempo_real_data}}
\subfigure[Performance comparison for unseen tasks.]{
\includegraphics[width=0.31\textwidth]{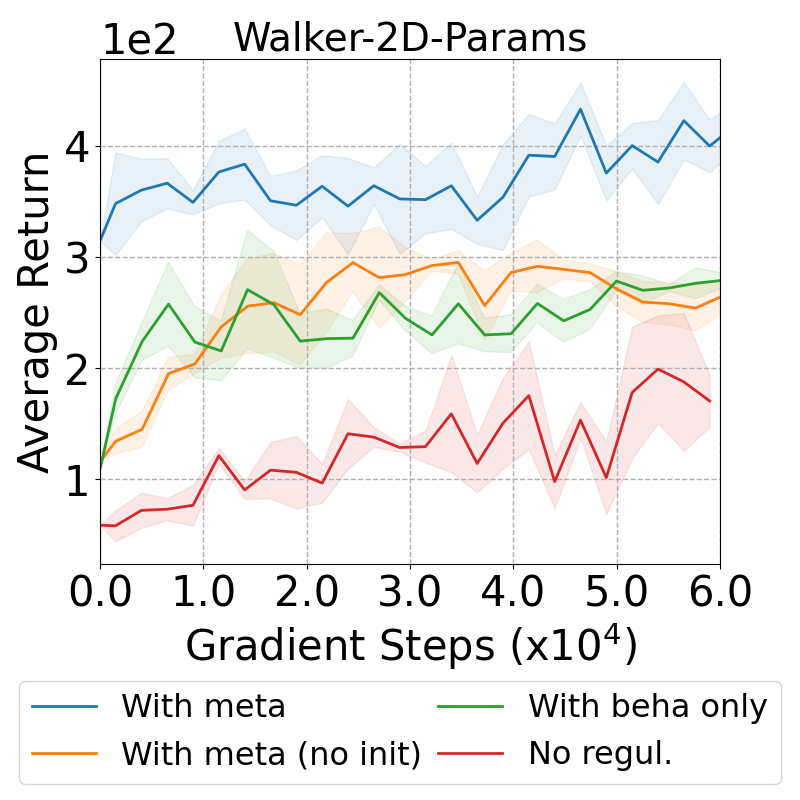}
\label{fig:rempo_test_only}}
\subfigure[Training sample efficiency in Point-Robot-Wind.]{
\includegraphics[width=0.31\textwidth]{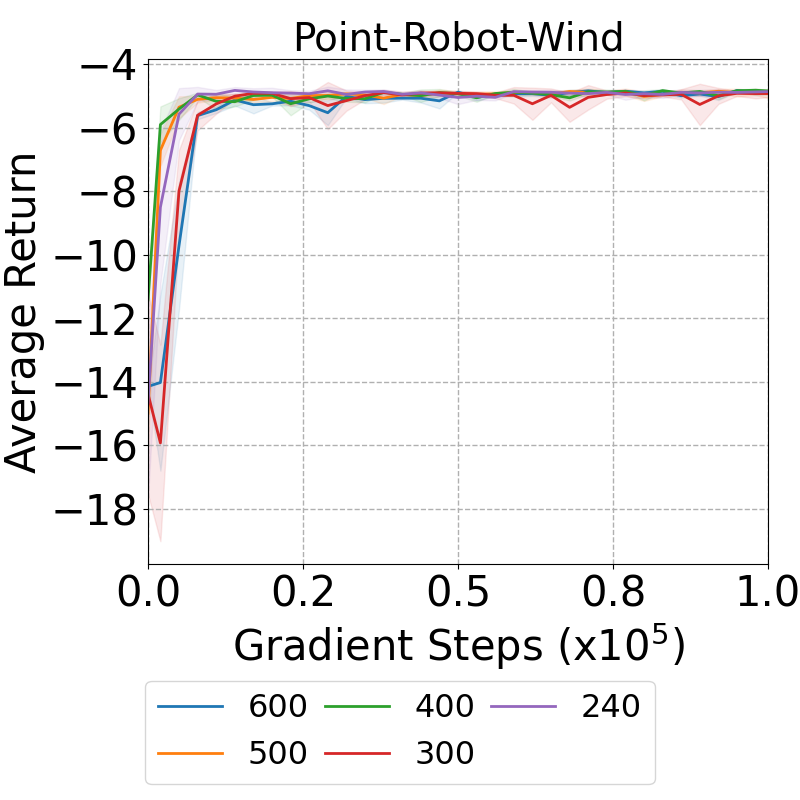}
\label{fig:rempo_effi_prw}}

\caption{Ablation study of MerPO.}
\label{fig:rempo_ablation}
\end{figure}

\subsubsection{More experiments.} 

We also evaluate the impact of the utilization extent of the learnt model, by comparing the performance of MerPO under different cases of real data ratio, i.e., the ratio of the data from the offline dataset in the data batch for training. As shown in Figure \ref{fig:rempo_real_data}, the performance of MerPO can be further boosted with a more conservative utilization of the model.

To understand how much benefit  MerPO can bring for policy learning in unseen offline RL tasks, we compare the performance of the following cases with respect to the gradient steps taken for learning in unseen offline RL tasks: (1) Initialize the task policy network with the meta-policy and run RAC (``With meta''); (2) Run RAC using the meta-policy without network initialization (``With meta (no init)''); (3) Run RAC with a single regularization based on behavior policy without network initialization, i.e., $\alpha=1$ (``With beha only''); (4) Run COMBO (``No regul.''). As shown in Figure \ref{fig:rempo_test_only}, ``With meta'' achieves the best performance and improves significantly over ``No regul.'' and ``With beha only'', i.e., learning alone without any guidance of meta-policy, which implies that the learnt meta-policy can efficiently guide the exploration of out-of-distribution state-actions. Without network initialization, ``With meta (no init)'' and ``With beha only'' achieve similar performance because good offline dataset is considered here. Such a result is also consistent with Figure \ref{fig:hf_exp_alpha}.

%\begin{wrapfigure}{r}{.53\textwidth}
% \vspace{-0.5cm}         
\begin{figure}
\centering  
\subfigure[Training efficiency.]{
\includegraphics[width=0.32\textwidth]{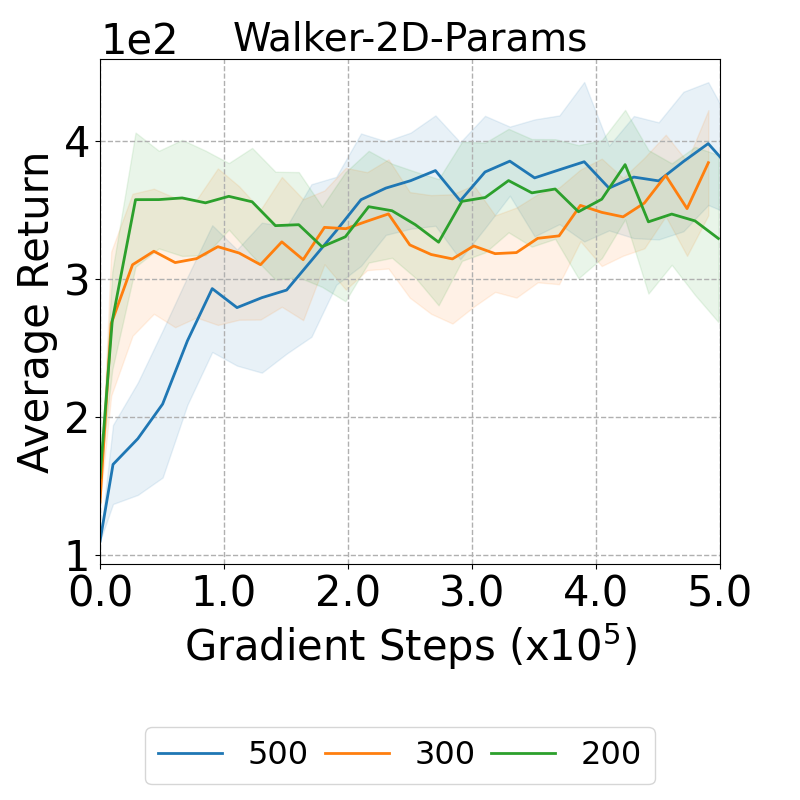}
\label{fig:rempo_effi_walker_new}}
\subfigure[Testing efficiency.]{
\includegraphics[width=0.32\textwidth]{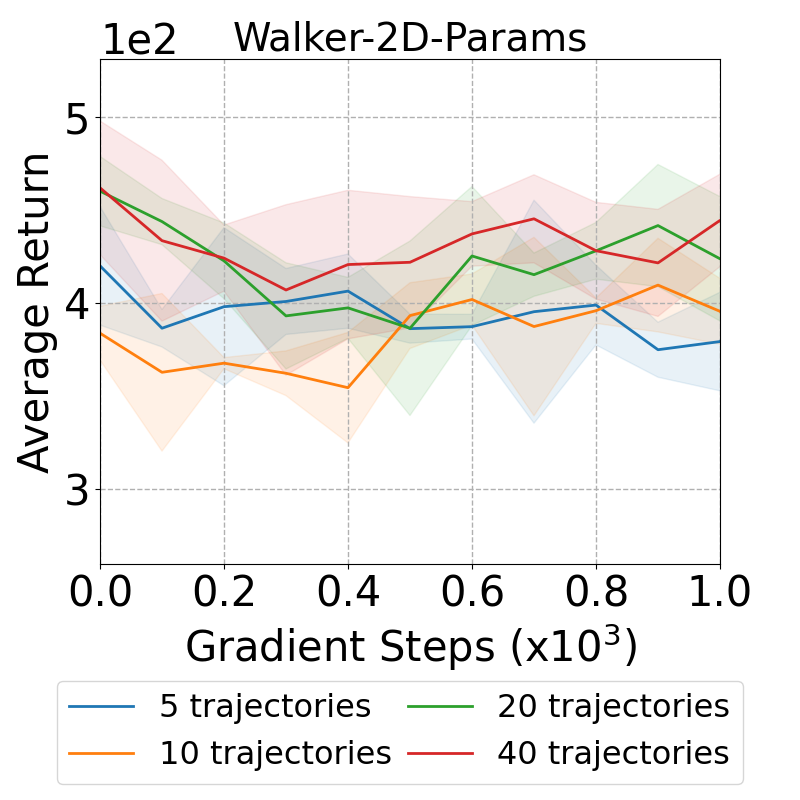}
\label{fig:rempo_effi_test_new}}
 \vspace{-0.3cm}
     \caption{Sample efficiency.}
         \label{fig:Merpo_sample}
% \vspace{-0.3cm}
\end{figure}
%        \vspace{-0.3cm}
%\end{wrapfigure}
%\textbf{(4) Is MerPO sample-efficient?} 
%Model-based RL approaches carry the promise to be more sample-efficient than model-free approaches. 

We evaluate the testing performance of MerPO, by changing sample size  of all tasks. Figure~\ref{fig:rempo_effi_walker_new} shows that the performance of MerPO is stable even if we decrease the number of trajectories for each task to be around 200. In contrast, the number of trajectories collected in other baselines is of the order $10^3$. Figure \ref{fig:rempo_effi_test_new} illustrates the testing sample efficiency of MerPO, by evaluating the performance at new offline tasks under different sample sizes. Clearly, a good task-specific policy can be quickly adapted at a new task even with 5 trajectories (1000 samples) of offline data. We also evaluate the training sample efficiency of MerPO in Point-Robot-Wind. As shown in Figure \ref{fig:rempo_effi_prw} the performance of MerPO is stable even if we decrease the number of trajectories for each task to be around 200. 

\subsubsection{More comparison between FOCAL and COMBO}

\begin{figure}
\centering
\subfigure[Performance comparison in Half-Cheetah-Fwd-Back.]{
\includegraphics[width=0.48\textwidth]{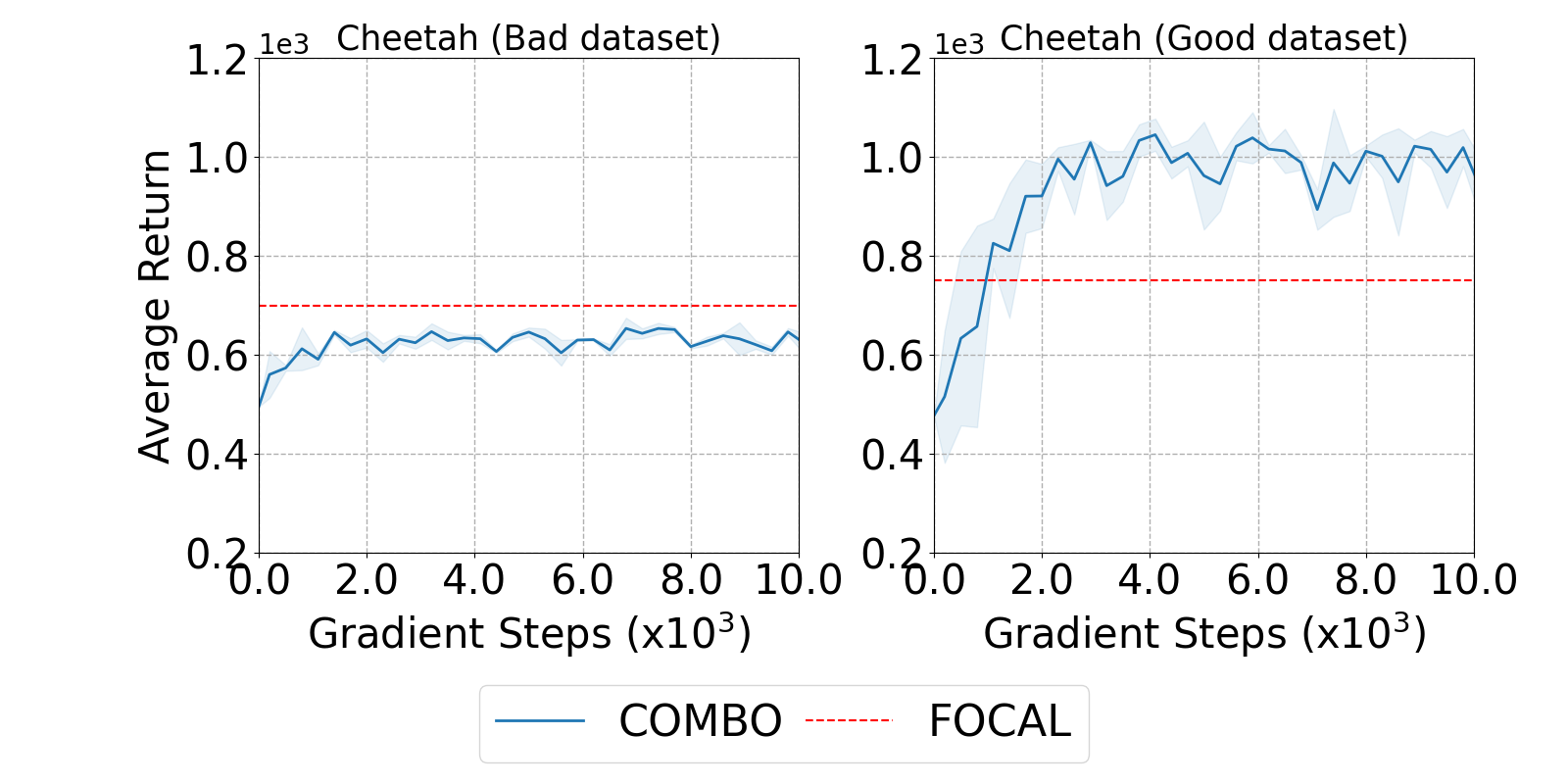}
\label{fig:hf}}
\subfigure[Performance comparison in Ant-Fwd-Back.]{
\includegraphics[width=0.48\textwidth]{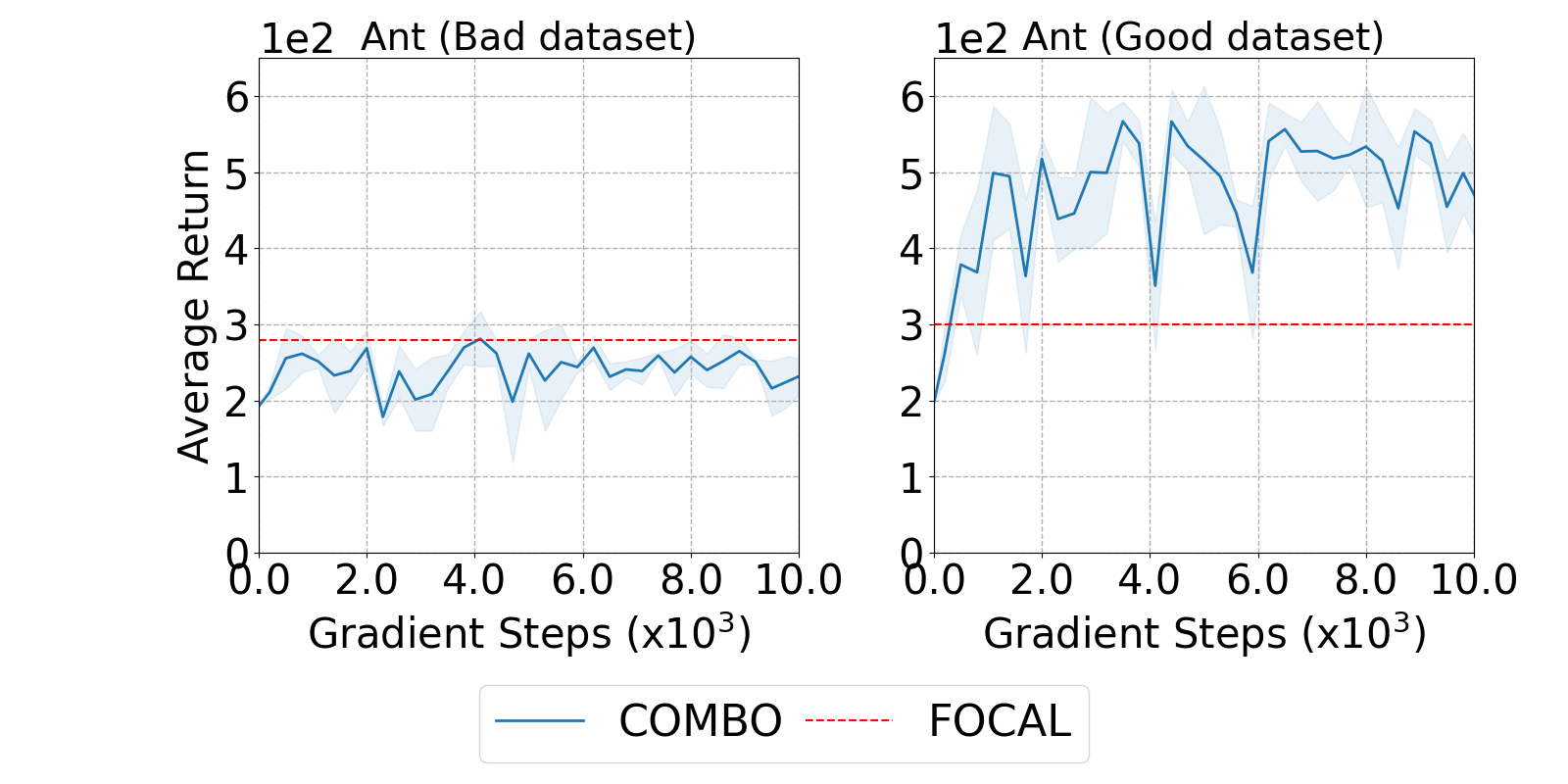}
\label{fig:ant}}
\caption{FOCAL vs. COMBO.}
\label{fig:intro_more}
\end{figure}

Following the setup as in Figure \ref{fig:intro}, we compare the performance between FOCAL and COMBO in two more environments: Half-Cheetah-Fwd-Back and Ant-Fwd-Back. As shown in Figure \ref{fig:intro_more}, although FOCAL performs better than COMBO on the task with a bad-quality dataset, it is outperformed by COMBO on the task with a good-quality dataset. This further confirms the observation made in Figure \ref{fig:intro}.

\subsection{Algorithms}

We include the details of MerPO in Algorithm \ref{alg2}.

\begin{algorithm}[tb]
	\caption{Regularized policy optimization for model-based offline Meta-RL (MerPO)}
	\label{alg2}
 	\begin{algorithmic}[1]
 	\State Initialize the dynamics, actor and critic for each task, and initialize the meta-model and the meta-policy;
 	\For{$k=1, 2, ...$}
 	    \For{each training task $\mathcal{M}_n$}
 	        \State Solve the following problem with gradient descent for $h$ steps to compute the dynamics model $\widehat{T}_{\theta^k_n}$ based on the offline dataset $\mathcal{D}_i$:
 	        \begin{align*}
 	            \min_{\theta_n}~~\mathbb{E}_{(s,a,s')\sim \mathcal{D}_n}[\log \widehat{T}_{\theta_n}(s'|s,a)]+\eta \|\theta_n-\phi_m(k)\|^2_2;
 	        \end{align*}
 	    \EndFor
 	    \State Update $\phi_m(k+1)=\phi_m(k)-\xi_1[\phi_m(k)-\frac{1}{N}\sum_{n=1}^N \theta^k_n]$;
 	 \EndFor
 	 \State Quickly obtain the estimated dynamics model $\widehat{T}_n$ for each training task by solving \cref{eq:model_local} with $t$ steps gradient descent;
	 \For{$k= 1, 2, ...$}
		    \For{each training task $\mathcal{M}_n$}
		        \For{$j=1,...,J$}
		            \State Perform model rollouts with $\widehat{T}_n$ starting from states in $\mathcal{D}_n$ and add model rollouts to $\mathcal{D}^n_{model}$;
		            \State Policy evaluation by recursively solving \cref{eq:pe_combo} using data from $\mathcal{D}_n\cup \mathcal{D}^n_{model}$;
		            \State Given the meta-policy $\pi_c^k$, improve policy $\pi_n^k$  by solving \cref{eq:pi_our};
		          \EndFor
		      %  \State Add model rollouts to $D^{model}_i$;
		      %  \State Conservatively evaluate current policy by minimizing the regularized critic loss $L^c_i$ using data from $D_i\cup D^{model}_i$:
		      %  \begin{align*}
		      %      \hat{Q}_i^{k+1}\leftarrow \arg\min_Q \beta(\mathbb{E}_{s,a\sim\rho_i(s,a)}[Q(s,a)]-\mathbb{E}_{s,a\sim D_i}[Q(s,a)])+\frac{1}{2}\mathbb{E}_{s,a,s'\sim d_{i,f}}\left[\left(Q(s,a)-\hat{\mathcal{B}}^{\pi}\hat{Q}_i^k(s,a)\right)^2\right]
		      %  \end{align*}
		       
		      %  \State Improve policy by minimizing the regularized actor loss $L^a_i$:
		      %  \begin{align*}
		      %      \pi_i'\leftarrow \arg\max_{\pi} \mathbb{E}_{s\sim \rho_i, a\sim \pi(\cdot|s)} \left[\hat{Q}_i^{\pi}(s,a)\right]-\lambda\alpha \mathbb{E}_{(s,a)\sim D} [\log \pi_{o}(a|s)]-\lambda (1-\alpha)D_{KL}(\pi_{c}||\pi)
		      %  \end{align*}
		    \EndFor
		   \State Given the learnt policy $\pi_n^k$ for each task, update the meta-policy $\pi_c^{k+1}$ by solving \cref{eq:meta-policy} with one step gradient descent;
		\EndFor
	\end{algorithmic}
\end{algorithm}

\section{Preliminaries}
\label{ap:pre}

For ease of exposition, let $T_{\mathcal{M}}$ and $r_{\mathcal{M}}$ denote the dynamics and reward function of the underlying MDP $\mathcal{M}$, $T_{\overline{\mathcal{M}}}$ and $r_{\overline{\mathcal{M}}}$ denote the dynamics and reward function of the empirical MDP $\overline{\mathcal{M}}$ induced by the dataset $\mathcal{D}$, and $T_{\widehat{\mathcal{M}}}$ and $r_{\widehat{\mathcal{M}}}$ denote the dynamics and reward function of the learnt MDP $\widehat{\mathcal{M}}$. To prevent any trivial bound with $\infty$ values, we assume that the cardinality of a state-action pair in the dataset $\mathcal{D}$, i.e., $|\mathcal{D}(s,a)|$, in the denominator, is non-zero, by setting $|\mathcal{D}(s,a)|$ to be a small value less than 1 when $(s,a)\notin \mathcal{D}$.

Following the same line as in \citep{kumar2020conservative,yu2021combo}, we make the following standard assumption on the concentration properties of the reward and dynamics for the empirical MDP $\overline{\mathcal{M}}$ to characterize the sampling error.

\begin{assumption}\label{assumption1}
For any $(s,a)\in \mathcal{M}$, the following inequalities hold with probability $1-\delta$:
\begin{align*}
    \|T_{\overline{\mathcal{M}}}(s'|s,a)-T_{\mathcal{M}}(s'|s,a)\|_1\leq \frac{C_{T,\delta}}{\sqrt{|\mathcal{D}(s,a)|}}; \ \ ~|r_{\overline{\mathcal{M}}}(s,a)-r_{\mathcal{M}}|\leq \frac{C_{r,\delta}}{\sqrt{|\mathcal{D}(s,a)|}}
\end{align*}
where $C_{T,\delta}$ and $C_{r,\delta}$ are some constants depending on $\delta$ via a $\sqrt{\log(1/\delta)}$ dependency.
\end{assumption}

Based on Assumption \ref{assumption1}, we can bound the estimation error induced by the empirical Bellman backup operator for any $(s,a)\in \mathcal{M}$:
\begingroup
\allowdisplaybreaks
\begin{align*}
    &\left|\mathcal{B}^{\pi}_{\overline{\mathcal{M}}}\widehat{Q}^k(s,a)-\mathcal{B}^{\pi}_{\mathcal{M}}\widehat{Q}^k(s,a)\right|\\
    =&\left|r_{\overline{\mathcal{M}}}(s,a)-r_{\mathcal{M}}(s,a)+\gamma \sum_{s'}(T_{\overline{\mathcal{M}}}(s'|s,a)-T_{\mathcal{M}}(s'|s,a))\mathbb{E}_{\pi(a'|s')}[\widehat{Q}^k(s',a')]\right|\\
    \leq& |r_{\overline{\mathcal{M}}}(s,a)-r_{\mathcal{M}}(s,a)|+\gamma\left|\sum_{s'}(T_{\overline{\mathcal{M}}}(s'|s,a)-T_{\mathcal{M}}(s'|s,a))\mathbb{E}_{\pi(a'|s')}[\widehat{Q}^k(s',a')]\right|\\
    \leq& \frac{C_{r,\delta}}{\sqrt{|\mathcal{D}(s,a)|}}+\gamma \|T_{\overline{\mathcal{M}}}(s'|s,a)-T_{\mathcal{M}}(s'|s,a)\|_1 \|\mathbb{E}_{\pi(a'|s')}[\widehat{Q}^k(s',a')]\|_{\infty}\\
    \leq&\frac{C_{r,\delta}+\gamma C_{T,\delta}R_{max}/(1-\gamma)}{\sqrt{|\mathcal{D}(s,a)|}}\\
    =& \frac{{((1-\gamma)C_{r,\delta}/R_{max}+\gamma C_{T,\delta})R_{max}}}{(1-\gamma)\sqrt{|\mathcal{D}(s,a)|}}\\
    \leq& \frac{{(C_{r,\delta}/R_{max}+ C_{T,\delta})R_{max}}}{(1-\gamma)\sqrt{|\mathcal{D}(s,a)|}}\triangleq \frac{C_{r,T,\delta}R_{max}}{(1-\gamma)\sqrt{|\mathcal{D}(s,a)|}}.
\end{align*}
\endgroup

Similarly, we can bound the difference between the Bellman backup induced by the learnt MDP $\widehat{\mathcal{M}}$ and the underlying Bellman backup:
\begin{align*}
    &\left|\mathcal{B}^{\pi}_{\widehat{\mathcal{M}}}\widehat{Q}^k(s,a)-\mathcal{B}^{\pi}_{\mathcal{M}}\widehat{Q}^k(s,a)\right|\\
    \leq& |r_{\widehat{\mathcal{M}}}(s,a)-r_{\mathcal{M}}(s,a)|+\frac{\gamma R_{max}}{1-\gamma}D_{tv}(T_{\widehat{\mathcal{M}}},T_{\mathcal{M}})
\end{align*}
where $D_{tv}(T_{\widehat{\mathcal{M}}},T_{\mathcal{M}})$ is the total-variation distance between $T_{\widehat{\mathcal{M}}}$ and $T_{\mathcal{M}}$.

For any two MDPs, $\mathcal{M}_1$ and $\mathcal{M}_2$, with the same state space, action space and discount factor $\gamma$, and a given fraction $f\in (0,1)$, define the $f$-interpolant MDP $\mathcal{M}_f$ as the MDP with dynamics: $T_{\mathcal{M}_f}=f T_{\mathcal{M}_1}+(1-f)T_{\mathcal{M}_2}$ and reward function: $r_{\mathcal{M}_f}=f r_{\mathcal{M}_1}+(1-f)r_{\mathcal{M}_2}$, which has the same state space, action space and discount factor with $\mathcal{M}_1$ and $\mathcal{M}_2$. Let $T^{\pi}$ be the transition matrix on state-action pairs induced by a stationary policy $\pi$, i.e.,
\begin{align*}
    T^{\pi}=T(s'|s,a)\pi(a'|s').
\end{align*}

To prove the main result, we first restate the following lemma from \citep{yu2021combo} to be used later.
\begin{lemma}\label{lemma1}
 For any policy $\pi$,  its returns in any MDP $\mathcal{M}$, denoted by $J(\mathcal{M},\pi)$, and in $\mathcal{M}_f$, denoted by $J(\mathcal{M}_1,\mathcal{M}_2, f,\pi)$, satisfy the following:
\begin{align*}
    J(\mathcal{M},\pi)-\eta \leq J(\mathcal{M}_1,\mathcal{M}_2,f,\pi)\leq J(\mathcal{M},\pi)+\eta
\end{align*}
where
\begin{align*}
    \eta =&\frac{2\gamma(1-f)}{(1-\gamma)^2} R_{max} D_{tv}(T_{\mathcal{M}_2}, T_{\mathcal{M}})+\frac{\gamma f}{1-\gamma} |\mathbb{E}_{d^{\pi}_{\mathcal{M}}} [(T^{\pi}_{\mathcal{M}}-T^{\pi}_{\mathcal{M}_1})Q_{\mathcal{M}}^{\pi}] |\\
	&+\frac{f}{1-\gamma} \mathbb{E}_{s,a\sim d^{\pi}_{\mathcal{M}}}[|r_{\mathcal{M}_1}(s,a)-r_{\mathcal{M}}(s,a)|]+\frac{1-f}{1-\gamma} \mathbb{E}_{s,a\sim d^{\pi}_M}[|r_{\mathcal{M}_2}(s,a)-r_{\mathcal{M}}(s,a)|].
\end{align*}
\end{lemma}

Lemma \ref{lemma1} characterizes the relationship between policy returns in different MDPs in terms of the corresponding reward difference and dynamics difference.

\section{Proof of Theorem 1}
\label{ap:proof}

Let $d(s,a):=d^{\pi_{\beta}}_{\mathcal{M}}(s,a)$.
In the setting without function approximation, by setting the derivation of Equation \cref{eq:pe_combo} to 0, we have that
\begin{align*}
    \widehat{Q}^{k+1}(s,a) = \widehat{\mathcal{B}}^{\pi}\widehat{Q}^k(s,a)-\beta \frac{\rho(s,a)-d(s,a)}{d_f(s,a)}.
\end{align*}
Denote $\nu(\rho,f)=\mathbb{E}_{\rho} \left[\frac{\rho(s,a)-d(s,a)}{d_f(s,a)}\right]$ as the expected penalty on the Q-value. It can be shown \citep{yu2021combo} that $\nu(\rho,f)\geq 0$ and increases with $f$, for any $\rho$ and $f\in (0,1)$.
Then, RAC optimizes the return of a policy in a $f$-interpolant MDP induced by the empirical MDP $\overline{\mathcal{M}}$ and the learnt MDP $\widehat{\mathcal{M}}$, which is regularized by both the behavior policy $\pi_{\beta}$ and the meta-policy $\pi_c$:
\begin{align}\label{op:underlying}
	\max_{\pi}~~ J(\overline{\mathcal{M}},\widehat{\mathcal{M}},f, \pi)-\beta \frac{\nu(\rho^{\pi},f)}{1-\gamma}-\lambda\alpha D(\pi,\pi_{\beta})-\lambda(1-\alpha) D(\pi,\pi_{c}).
\end{align}

Denote $\pi_o$ as the solution to the above optimization problem. Based on Lemma \ref{lemma1}, we can first characterize the return of the learnt policy $\pi_o$ in the underlying MDP $\mathcal{M}$ in terms of its return in the $f$-interpolant MDP:
\begin{align}\label{eq:lowerbound}
	J(M, \pi_o)+\eta_1\geq J(\overline{\mathcal{M}},\widehat{\mathcal{M}},f, \pi_o)
\end{align}
where 
\begin{align*}
	\eta_1=&\frac{2\gamma(1-f)}{(1-\gamma)^2} R_{max} D_{tv}(T_{\widehat{\mathcal{M}}}, T_{\mathcal{M}})+\frac{\gamma f}{1-\gamma} |\mathbb{E}_{d^{\pi_o}_{\mathcal{M}}} [(T^{\pi_o}_{\mathcal{M}}-T^{\pi_o}_{\overline{\mathcal{M}}})Q_{\mathcal{M}}^{\pi_o}] |\\
	&+\frac{f}{1-\gamma} \mathbb{E}_{s,a\sim d^{\pi_o}_{\mathcal{M}}}[|r_{\overline{\mathcal{M}}}(s,a)-r_{\mathcal{M}}(s,a)|]+\frac{1-f}{1-\gamma} \mathbb{E}_{s,a\sim d^{\pi_o}_{\mathcal{M}}}[|r_{\widehat{\mathcal{M}}}(s,a)-r_{\mathcal{M}}(s,a)|]\\
	\leq & \frac{2\gamma (1-f)}{(1-\gamma)^2} R_{max} D_{tv}(T_{\widehat{\mathcal{M}}}, T_{\mathcal{M}})+\frac{\gamma^2 f C_{T,\delta}R_{max}}{(1-\gamma)^2}\mathbb{E}_{s\sim d^{\pi_o}_{\mathcal{M}} (s)}\left[\sqrt{\frac{|A|}{|\mathcal{D}(s)|}}\sqrt{D_{CQL}(\pi_o,\pi_{\beta})(s)+1}\right]\\
	&+\frac{C_{r,\delta}}{1-\gamma} \mathbb{E}_{s,a\sim d_{\mathcal{M}}^{\pi_o}}\left[\frac{1}{\sqrt{|\mathcal{D}(s,a)}|}\right]+\frac{1}{1-\gamma}\mathbb{E}_{s,a\sim d^{\pi_o}_{\mathcal{M}}}[|r_{\widehat{\mathcal{M}}}(s,a)-r_{\mathcal{M}}(s,a)|]\\
	\triangleq & \eta_1^c.
\end{align*}
Note that the inequality above holds because the following is true for the empirical MDP $\overline{\mathcal{M}}$  \citep{kumar2020conservative}:
\begin{align*}
    |\mathbb{E}_{d^{\pi}_{\mathcal{M}}} [(T^{\pi}_{\mathcal{M}}-T^{\pi}_{\overline{\mathcal{M}}})Q_{\mathcal{M}}^{\pi}] |\leq \frac{\gamma C_{T,\delta} R_{max}}{1-\gamma}\mathbb{E}_{s\sim d^{\pi}_{\mathcal{M}} (s)}\left[\sqrt{\frac{|A|}{|\mathcal{D}(s)|}}\sqrt{D_{CQL}(\pi,\pi_{\beta})(s)+1}\right]
\end{align*}
for $D_{CQL}(\pi_1,\pi_2)(s):=\sum_a \pi_1(a|s)\left(\frac{\pi_1(a|s)}{\pi_2(a|s)}-1\right)$.

\subsection{Safe improvement over $\pi_c$}
We first show that the learnt policy offers safe improvement over the meta-policy $\pi_c$.
Following the same line as in \cref{eq:lowerbound}, we  next bound the return of the meta-policy $\pi_c$ in the underlying MDP $\mathcal{M}$ from above, in terms of its return in the $f$-interpolant MDP:
\begin{align*}
	J(\overline{\mathcal{M}},\widehat{\mathcal{M}},f, \pi_c)\geq J(M, \pi_c)-\eta_2
\end{align*}
where
\begin{align*}
	\eta_2 \leq& \frac{2\gamma (1-f)}{(1-\gamma)^2} R_{max} D_{tv}(T_{\widehat{\mathcal{M}}}, T_{\mathcal{M}})+\frac{\gamma^2 f C_{T,\delta}R_{max}}{(1-\gamma)^2}\mathbb{E}_{s\sim d^{\pi_c}_{\mathcal{M}} (s)}\left[\sqrt{\frac{|A|}{|\mathcal{D}(s)|}}\sqrt{D_{CQL}(\pi_c,\pi_{\beta})(s)+1}\right]\\
	&+\frac{C_{r,\delta}}{1-\gamma} \mathbb{E}_{s,a\sim d_{\mathcal{M}}^{\pi_c}}\left[\frac{1}{\sqrt{|\mathcal{D}(s,a)|}}\right]+\frac{1}{1-\gamma}\mathbb{E}_{s,a\sim d^{\pi_c}_{\mathcal{M}}}[|r_{\widehat{\mathcal{M}}}(s,a)-r_{\mathcal{M}}(s,a)|]\\
	\triangleq & \eta_2^c.
\end{align*}

It follows that
\begin{align*}
	&J(\mathcal{M},\pi_o)+\eta_1^c-\beta \frac{\nu(\rho^{\pi_o},f)}{1-\gamma}-\lambda\alpha D(\pi_o,\pi_{\beta})-\lambda(1-\alpha) D(\pi_o,\pi_{c})\\
	\geq & J(\overline{\mathcal{M}},\widehat{\mathcal{M}},f,\pi_o)-\beta \frac{\nu(\rho^{\pi_o},f)}{1-\gamma}-\lambda\alpha D(\pi_o,\pi_{\beta})-\lambda(1-\alpha) D(\pi_o,\pi_{c})\\
	\geq & J(\overline{\mathcal{M}},\widehat{\mathcal{M}},f,\pi_c)-\beta \frac{\nu(\rho^{\pi_c},f)}{1-\gamma}-\lambda\alpha D(\pi_c,\pi_{\beta})\\
	\geq & J(\mathcal{M}, \pi_c)-\eta_2^c-\beta \frac{\nu(\rho^{\pi_c},f)}{1-\gamma}-\lambda\alpha D(\pi_c,\pi_{\beta}),
\end{align*}
where the second inequality is true because $\pi_o$ is the solution to \cref{op:underlying}. This gives us a lower bound on $J(\mathcal{M},\pi_o)$ in terms of $J(\mathcal{M},\pi_c)$:
\begin{align*}
	J(\mathcal{M}, \pi_o)
	\geq J(\mathcal{M}, \pi_c)-&\eta_1^c-\eta_2^c+\frac{\beta}{1-\gamma}[\nu(\rho^{\pi_o},f)-\nu(\rho^{\pi_c},f)]\\
	&+\lambda\alpha D(\pi_o,\pi_{\beta})+\lambda (1-\alpha) D(\pi_o,\pi_c)-\lambda\alpha D(\pi_c,\pi_{\beta}).
\end{align*}

It is clear that $\eta_1^c$ and $\eta_2^c$ are independent to $\beta$ and $\lambda$. To show the performance improvement of $\pi_o$ over the meta-policy $\pi_c$, it suffices to guarantee that for appropriate choices of $\beta$ and $\lambda$,
\begin{align*}
	\Delta_c=\lambda\alpha D(\pi_o,\pi_{\beta})+\lambda (1-\alpha) D(\pi_o,\pi_c)-\lambda\alpha D(\pi_c,\pi_{\beta})+\frac{\beta}{1-\gamma}[\nu(\rho^{\pi_o},f)-\nu(\rho^{\pi_c},f)]>0.
\end{align*}

To this end, the following lemma first provides an upper bound on $|\nu(\rho^{\pi_o},f)-\nu(\rho^{\pi_c},f)|$:

\begin{lemma}\label{lemma2}
%Let $D(\pi_1||\pi_2)=\max_s D_{tv}(\pi_1||\pi_2)$ denote the maximum total-variation distance between two policies $\pi_1$ and $\pi_2$. Then 
There exist some positive constants $L_1$ and $L_2$ such that
\begin{align*}
    |\nu(\rho^{\pi_o},f)-\nu(\rho^{\pi_c},f)|\leq 2(L_1+L_2) D_{tv}(\rho^{\pi_{o}}(s,a)||\rho^{\pi_{c}}(s,a)).
\end{align*}
\end{lemma}

\begin{proof}
First, we  have that
\begingroup
\allowdisplaybreaks
\begin{align*}
    & |\nu(\rho^{\pi_o},f)-\nu(\rho^{\pi_c},f)|\\
    =& \left|\mathbb{E}_{\rho^{\pi_{o}}}\left[\frac{\rho^{\pi_o}(s,a)-d(s,a)}{f d(s,a)+(1-f)\rho^{\pi_o}(s,a)}\right]-\mathbb{E}_{\rho^{\pi_{c}}}\left[\frac{\rho^{\pi_c}(s,a)-d(s,a)}{f d(s,a)+(1-f)\rho^{\pi_c}(s,a)}\right]\right|\\
    =&\left|\sum_{(s,a)} \left[\rho^{\pi_{o}}(s,a)\frac{\rho^{\pi_o}(s,a)-d(s,a)}{f d(s,a)+(1-f)\rho^{\pi_o}(s,a)}-\rho^{\pi_{c}}(s,a)\frac{\rho^{\pi_c}(s,a)-d(s,a)}{f d(s,a)+(1-f)\rho^{\pi_c}(s,a)}\right]\right|\\
    \leq&\left|\sum_{(s,a)} \left[\rho^{\pi_{o}}(s,a)\frac{\rho^{\pi_o}(s,a)-d(s,a)}{f d(s,a)+(1-f)\rho^{\pi_o}(s,a)}-\rho^{\pi_{c}}(s,a)\frac{\rho^{\pi_o}(s,a)-d(s,a)}{f d(s,a)+(1-f)\rho^{\pi_o}(s,a)}\right]\right| \\
    &+\left|\sum_{(s,a)}\left[\rho^{\pi_{c}}(s,a)\frac{\rho^{\pi_o}(s,a)-d(s,a)}{f d(s,a)+(1-f)\rho^{\pi_o}(s,a)}-\rho^{\pi_{c}}(s,a)\frac{\rho^{\pi_c}(s,a)-d(s,a)}{f d(s,a)+(1-f)\rho^{\pi_c}(s,a)}\right]\right|\\
    =&\left|\sum_{(s,a)}[\rho^{\pi_{o}}(s,a)-\rho^{\pi_{c}}(s,a)]\frac{\rho^{\pi_o}(s,a)-d(s,a)}{f d(s,a)+(1-f)\rho^{\pi_o}(s,a)}\right|\\
    &+\left|\sum_{(s,a)}\rho^{\pi_{c}}(s,a)\left[\frac{\rho^{\pi_o}(s,a)-d(s,a)}{f d(s,a)+(1-f)\rho^{\pi_o}(s,a)}-\frac{\rho^{\pi_c}(s,a)-d(s,a)}{f d(s,a)+(1-f)\rho^{\pi_c}(s,a)}\right]\right|\\
    \leq& \sum_{(s,a)}|\rho^{\pi_{o}}(s,a)-\rho^{\pi_{c}}(s,a)|\left|\frac{\rho^{\pi_o}(s,a)-d(s,a)}{f d(s,a)+(1-f)\rho^{\pi_o}(s,a)}\right|\\
    +&\sum_{(s,a)}\rho^{\pi_{c}}(s,a)\left|\frac{\rho^{\pi_o}(s,a)-d(s,a)}{f d(s,a)+(1-f)\rho^{\pi_o}(s,a)}-\frac{\rho^{\pi_c}(s,a)-d(s,a)}{f d(s,a)+(1-f)\rho^{\pi_c}(s,a)}\right|.
\end{align*}
\endgroup

First, observe that for the term $\left|\frac{\rho^{\pi_o}(s,a)-d(s,a)}{f d(s,a)+(1-f)\rho^{\pi_o}(s,a)}\right|$,  
\begin{itemize}
    \item If $\rho^{\pi_o}(s,a)\geq d(s,a)$, then
    \begin{align*}
        &\left|\frac{\rho^{\pi_o}(s,a)-d(s,a)}{f d(s,a)+(1-f)\rho^{\pi_o}(s,a)}\right|\\
        \leq& \left|\frac{\rho^{\pi_o}(s,a)}{f d(s,a)+(1-f)\rho^{\pi_o}(s,a)}\right|\leq \left|\frac{\rho^{\pi_o}(s,a)}{(1-f)\rho^{\pi_o}(s,a)}\right|=\frac{1}{1-f}.
    \end{align*}    
    \item If $\rho^{\pi_o}(s,a)< d(s,a)$, then
    \begin{align*}
        \left|\frac{\rho^{\pi_o}(s,a)-d(s,a)}{f d(s,a)+(1-f)\rho^{\pi_o}(s,a)}\right|\leq \left|\frac{d(s,a)}{f d(s,a)+(1-f)\rho^{\pi_o}(s,a)}\right|\leq \left|\frac{d(s,a)}{f d(s,a)}\right|=\frac{1}{f}.
    \end{align*}
\end{itemize}
Therefore, 
\begin{align*}
    \left|\frac{\rho^{\pi_o}(s,a)-d(s,a)}{f d(s,a)+(1-f)\rho^{\pi_o}(s,a)}\right|\leq \max\left\{\frac{1}{f},\frac{1}{1-f}\right\}\triangleq L_1.
\end{align*}

Next, for the term $\left|\frac{\rho^{\pi_o}(s,a)-d(s,a)}{f d(s,a)+(1-f)\rho^{\pi_o}(s,a)}-\frac{\rho^{\pi_c}(s,a)-d(s,a)}{f d(s,a)+(1-f)\rho^{\pi_c}(s,a)}\right|$, consider the function $g(x)=\frac{x-d}{fd+(1-f)x}$ for $x\in [0,1]$. Clearly, when $d(s,a)=0$,
\begin{align*}
    \left|\frac{\rho^{\pi_o}(s,a)-d(s,a)}{f d(s,a)+(1-f)\rho^{\pi_o}(s,a)}-\frac{\rho^{\pi_c}(s,a)-d(s,a)}{f d(s,a)+(1-f)\rho^{\pi_c}(s,a)}\right|=0.
\end{align*}
For any $(s,a)$ that $d(s,a)>0$, it can be shown that $g(x)$ is continuous and has bounded gradient, i.e., $|\nabla g(x)|\leq \frac{1}{f^2 d}\triangleq L_2$. Hence, it follows that 
\begin{align*}
    \left|\frac{\rho^{\pi_o}(s,a)-d(s,a)}{f d(s,a)+(1-f)\rho^{\pi_o}(s,a)}-\frac{\rho^{\pi_c}(s,a)-d(s,a)}{f d(s,a)+(1-f)\rho^{\pi_c}(s,a)}\right|\leq L_2 |\rho^{\pi_o}(s,a)-\rho^{\pi_c}(s,a)|.
\end{align*}

Therefore, we can conclude that
\begin{align*}
    & |\nu(\rho^{\pi_o},f)-\nu(\rho^{\pi_c},f)|\\
    \leq & L_1\sum_{(s,a)}|\rho^{\pi_{o}}(s,a)-\rho^{\pi_{c}}(s,a)|+L_2\sum_{(s,a)} \rho^{\pi_{c}}(s,a)|\rho^{\pi_{o}}(s,a)-\rho^{\pi_{c}}(s,a)|\\
    \leq & (L_1+L_2)\sum_{s,a}|\rho^{\pi_{o}}(s,a)-\rho^{\pi_{c}}(s,a)|\\
    =& 2(L_1+L_2) D_{tv}(\rho^{\pi_{o}}(s,a)||\rho^{\pi_{c}}(s,a)).
\end{align*}

\end{proof}

Recall that
\begin{align*}
    \rho^{\pi_{o}}(s,a)=d_{\widehat{\mathcal{M}}}^{\pi_o}(s)\pi_o(a|s),~~\rho^{\pi_{c}}(s,a)=d_{\widehat{\mathcal{M}}}^{\pi_c}(s)\pi_c(a|s),
\end{align*}
which denote the marginal state-action distributions by rolling out $\pi_o$ and $\pi_c$ in the learnt model $\widehat{\mathcal{M}}$, respectively. Lemma \ref{lemma2} gives an upper bound on the difference between the expected penalties induced under $\pi_o$ and $\pi_c$, with regard to the difference between the marginal state-action distributions. Next, we need to characterize the relationship between the marginal state-action distribution difference and the corresponding policy distance, which is captured in the following lemma.

\begin{lemma}\label{lemma3}
Let $D(\pi_1||\pi_2)=\max_s D_{tv}(\pi_1||\pi_2)$ denote the maximum total-variation distance between two policies $\pi_1$ and $\pi_2$. Then, we can have that
\begin{align*}
    D_{tv}(\rho^{\pi_{o}}(s,a)||\rho^{\pi_{c}}(s,a))\leq \frac{1}{1-\gamma}\max_s D_{tv}(\pi_o(a|s)||\pi_c(a|s)).
\end{align*}
\end{lemma}

\begin{proof}
Note that
\begin{align*}
    D_{tv}(\rho^{\pi_{o}}(s,a)||\rho^{\pi_{c}}(s,a))\leq (1-\gamma)\sum_{t=0}^{\infty} \gamma^t D_{tv}(\rho_t^{\pi_{o}}(s,a)||\rho_t^{\pi_{c}}(s,a)).
\end{align*}
It then suffices to bound the state-action marginal difference at time $t$. Since both state-action marginals here correspond to rolling out $\pi_o$ and $\pi_c$ in the same MDP $\widehat{\mathcal{M}}$, based on Lemma B.1 and B.2 in \citep{janner2019trust}, we can obtain that
\begin{align*}
    &D_{tv}(\rho_t^{\pi_{o}}(s,a)||\rho_t^{\pi_{c}}(s,a))\\
    \leq& D_{tv}(\rho_t^{\pi_{o}}(s)||\rho_t^{\pi_{c}}(s))+\max_s D_{tv} (\pi_o(a|s)||\pi_c(a|s))\\
    \leq& t\max_s D_{tv} (\pi_o(a|s)||\pi_c(a|s))+\max_s D_{tv} (\pi_o(a|s)||\pi_c(a|s))\\
    =& (t+1)\max_s D_{tv} (\pi_o(a|s)||\pi_c(a|s)),
\end{align*}
which indicates that
\begin{align*}
    D_{tv}(\rho^{\pi_{o}}(s,a)||\rho^{\pi_{c}}(s,a))&\leq (1-\gamma) \sum_{t=0}^{\infty} \gamma^t (t+1) \max_s D_{tv}(\pi_o(a|s)||\pi_c(a|s))\\
    &=\frac{1}{1-\gamma}\max_s D_{tv}(\pi_o(a|s)||\pi_c(a|s)).
\end{align*}
\end{proof}

Building on Lemma \ref{lemma2} and Lemma \ref{lemma3}, we can show that
\begin{align*}
    |\nu(\rho^{\pi_o},f)-\nu(\rho^{\pi_c},f)|&\leq \frac{2(L_1+L_2)}{1-\gamma} \max_s D_{tv}(\pi_o(a|s)||\pi_c(a|s))\\
    &\triangleq C\max_s D_{tv}(\pi_o(a|s)||\pi_c(a|s)).
\end{align*}
Let $D(\cdot,\cdot)=\max_s D_{tv}(\cdot||\cdot)$. It is clear that for $\lambda\geq \lambda_0$ where $\lambda_0>\frac{C\beta}{(1-\gamma)(1-2\alpha)}$ and $\alpha< \frac{1}{2}$, 
\begin{align*}
	\Delta_c=&\lambda\alpha D(\pi_o,\pi_{\beta})+\lambda (1-\alpha) D(\pi_o,\pi_c)-\lambda\alpha D(\pi_c,\pi_{\beta})+\frac{\beta}{1-\gamma}[\nu(\rho^{\pi_o},f)-\nu(\rho^{\pi_c},f)]\\
	=& \lambda\alpha D(\pi_o,\pi_{\beta})+\lambda\alpha D(\pi_o,\pi_{c})-\lambda\alpha D(\pi_c,\pi_{\beta})
	+\lambda (1-2\alpha)D(\pi_o,\pi_{c})
	\\
	&+\frac{\beta}{1-\gamma}[\nu(\rho^{\pi_o},f)-\nu(\rho^{\pi_c},f)]\\
	\geq &\lambda (1-2\alpha)D(\pi_o,\pi_{c})+\frac{\beta}{1-\gamma}[\nu(\rho^{\pi_o},f)-\nu(\rho^{\pi_c},f)]\\
	\geq & \lambda (1-2\alpha)D(\pi_o,\pi_c)-\frac{C\beta}{1-\gamma}D(\pi_o,\pi_c)\\
	=& (\lambda-\lambda_0)(1-2\alpha)D(\pi_o,\pi_c)+\left[\lambda_0(1-2\alpha)-\frac{C\beta}{1-\gamma}\right]D(\pi_o,\pi_c)> 0.
\end{align*}

In a nutshell, we can conclude that with probability $1-\delta$
\begin{align*}
    J(\mathcal{M}, \pi_o)
	\geq J(\mathcal{M}, \pi_c)\underbrace{-\eta_1^c-\eta_2^c}_{(a)}\underbrace{+(\lambda-\lambda_0)(1-2\alpha)D(\pi_o,\pi_c)}_{(b)}\underbrace{+\left[\lambda_0(1-2\alpha)-\frac{C\beta}{1-\gamma}\right]D(\pi_o,\pi_c)}_{(c)},
\end{align*}
where (a) depends on $\delta$ but is independent to $\lambda$, (b) is positive and increases with $\lambda$, and (c) is positive. This implies that an appropriate choice of $\lambda$ will make term (b) large enough  to counteract term (a) and lead to the performance improvement over the meta-policy $\pi_c$:
\begin{align*}
    J(\mathcal{M}, \pi_o)
	\geq J(\mathcal{M}, \pi_c)+\xi_1
\end{align*}
where $\xi_1\geq 0$.

\subsection{Safe improvement over $\pi_{\beta}$}

Next, we show that the learnt policy $\pi_o$ achieves safe improvement over the behavior policy $\pi_{\beta}$. Based on Lemma \ref{lemma1}, we have
\begin{align*}
    J(\mathcal{M}_1,\mathcal{M}_2,f,\pi_{\beta})\geq J(\mathcal{M},\pi_{\beta})-\eta_3
\end{align*}
where 
\begin{align*}
    \eta_3=&\frac{2\gamma(1-f)}{(1-\gamma)^2} R_{max} D_{tv}(T_{\widehat{\mathcal{M}}}, T_{\mathcal{M}})+\frac{\gamma f}{1-\gamma} |\mathbb{E}_{d^{\pi_{\beta}}_{\mathcal{M}}} [(T^{\pi_{\beta}}_{\mathcal{M}}-T^{\pi_{\beta}}_{\overline{\mathcal{M}}})Q_{\mathcal{M}}^{\pi_{\beta}}] |\\
	&+\frac{f}{1-\gamma} \mathbb{E}_{s,a\sim d^{\pi_{\beta}}_{\mathcal{M}}}[|r_{\overline{\mathcal{M}}}(s,a)-r_{\mathcal{M}}(s,a)|]+\frac{1-f}{1-\gamma} \mathbb{E}_{s,a\sim d^{\pi_{\beta}}_{\mathcal{M}}}[|r_{\widehat{\mathcal{M}}}(s,a)-r_{\mathcal{M}}(s,a)|]\\
	\leq & \frac{2\gamma (1-f)}{(1-\gamma)^2} R_{max} D_{tv}(T_{\widehat{\mathcal{M}}}, T_{\mathcal{M}})+\frac{\gamma^2 f C_{T,\delta}R_{max}}{(1-\gamma)^2}\mathbb{E}_{s\sim d^{\pi_{\beta}}_{\mathcal{M}} (s)}\left[\sqrt{\frac{|A|}{|\mathcal{D}(s)|}}\right]\\
	&+\frac{C_{r,\delta}}{1-\gamma} \mathbb{E}_{s,a\sim d_{\mathcal{M}}^{\pi_{\beta}}}\left[\frac{1}{\sqrt{|\mathcal{D}(s,a)|}}\right]+\frac{1}{1-\gamma}\mathbb{E}_{s,a\sim d^{\pi_{\beta}}_{\mathcal{M}}}[|r_{\widehat{\mathcal{M}}}(s,a)-r_{\mathcal{M}}(s,a)|]\\
	\triangleq & \eta_3^{\beta}.
\end{align*}

Therefore, it follows that
\begin{align*}
	&J(\mathcal{M},\pi_o)+\eta_1^c-\beta \frac{\nu(\rho^{\pi_o},f)}{1-\gamma}-\lambda\alpha D(\pi_o,\pi_{\beta})-\lambda(1-\alpha) D(\pi_o,\pi_{c})\\
	\geq & J(\overline{\mathcal{M}},\widehat{\mathcal{M}},f,\pi_o)-\beta \frac{\nu(\rho^{\pi_o},f)}{1-\gamma}-\lambda\alpha D(\pi_o,\pi_{\beta})-\lambda(1-\alpha) D(\pi_o,\pi_{c})\\
	\geq & J(\overline{\mathcal{M}},\widehat{\mathcal{M}},f,\pi_{\beta})-\beta \frac{\nu(\rho^{\pi_{\beta}},f)}{1-\gamma}-\lambda(1-\alpha) D(\pi_{\beta},\pi_c)\\
	\geq & J(\mathcal{M}, \pi_{\beta})-\eta_3^{\beta}-\beta \frac{\nu(\rho^{\pi_{\beta}},f)}{1-\gamma}-\lambda(1-\alpha) D(\pi_{\beta},\pi_c),
\end{align*}
which indicates that with probability $1-\delta$
\begin{align*}
	J(\mathcal{M},\pi_c)\geq J(\mathcal{M},\pi_{\beta})-\eta_1^{c}-\eta_3^{\beta}+&\lambda\alpha D(\pi_o,\pi_{\beta})+\lambda (1-\alpha) D(\pi_o,\pi_c)-\lambda(1-\alpha) D(\pi_{\beta},\pi_c)\\&+\frac{\beta}{1-\gamma}[\nu(\rho^{\pi_o},f)-\nu(\rho^{\pi_{\beta}},f)],
\end{align*}
where $\eta_3^{\beta}$ is some constant that depends on $\delta$ but is independent to $\beta$ and $\lambda$.

To conclude, we can have that with probability $1-2\delta$
\begin{align*}
    J(\mathcal{M},\pi_o)\geq \max\{J(\mathcal{M},\pi_c)+\xi_1, J(\mathcal{M},\pi_{\beta})+\xi_2\}
\end{align*}
where 
\begin{align}\label{eq:xi1}
    \xi_1=-\eta_1^c-\eta_2^c+(\lambda-\lambda_0)(1-2\alpha)D(\pi_o,\pi_c)+\left[\lambda_0(1-2\alpha)-\frac{C\beta}{1-\gamma}\right]D(\pi_o,\pi_c)
\end{align}
and
\begin{align}\label{eq:xi2}
    \xi_2=-\eta_1^{c}-\eta_3^{\beta}+\lambda\alpha D(\pi_o,\pi_{\beta})+&\lambda (1-\alpha) D(\pi_o,\pi_c)-\lambda(1-\alpha) D(\pi_{\beta},\pi_c)\\
    &+\frac{\beta}{1-\gamma}[\nu(\rho^{\pi_o},f)-\nu(\rho^{\pi_{\beta}},f)].
\end{align}

Moreover, as we noted earlier, $\xi_1>0$ for a suitably selected $\lambda$ and $\alpha<\frac{1}{2}$. For the term $\nu(\rho^{\pi_o},f)-\nu(\rho^{\pi_{\beta}},f)$ in $\xi_2$ where $\nu(\rho^{\pi},f)$ is defined as $\mathbb{E}_{\rho^{\pi}} \left[\frac{\rho^{\pi}(s,a)-d(s,a)}{d_f(s,a)}\right]$, as noted in \citep{yu2021combo}, $\nu(\rho^{\pi_{\beta}},f)$ is expected to be smaller than $\nu(\rho^{\pi_o},f)$ in practical scenarios, due to the fact that the dynamics $T_{\widehat{\mathcal{M}}}$ learnt via supervised learning is close to the underlying dynamics $T_{\mathcal{M}}$ on the states visited by the behavior policy $\pi_{\beta}$. This directly indicates that $d^{\pi_{\beta}}_{\widehat{\mathcal{M}}}(s,a)$ is close to $d^{\pi_{\beta}}_{\overline{\mathcal{M}}}(s,a)$ and $\rho^{\pi_{\beta}}$ is close to $d(s,a)$. In this case, let
\begin{align*}
	\epsilon=\frac{\beta[\nu(\rho^{\pi_o},f)-\nu(\rho^{\pi_{\beta}},f)]}{2\lambda(1-\gamma) D(\pi_o,\pi_{\beta})}.
\end{align*}
We can show that for $\alpha>\frac{1}{2}-\epsilon$,
\begin{align*}
    \Delta_{\beta}=
    &\lambda\alpha D(\pi_o,\pi_{\beta})+\lambda (1-\alpha) D(\pi_o,\pi_c)-\lambda(1-\alpha) D(\pi_c,\pi_{\beta})+\frac{\beta}{1-\gamma}[\nu(\rho^{\pi_o},f)-\nu(\rho^{\pi_{\beta}},f)]\\
    =&\lambda\alpha D(\pi_o,\pi_{\beta})+\lambda (1-\alpha) D(\pi_o,\pi_c)-\lambda(1-\alpha) D(\pi_c,\pi_{\beta})+2\epsilon\lambda D(\pi_o,\pi_{\beta})\\
    =&\lambda\left[(2\epsilon+\alpha)D(\pi_o,\pi_{\beta})+(1-\alpha) D(\pi_o,\pi_c)-(1-\alpha) D(\pi_c,\pi_{\beta})\right]\\
    >& \lambda(1-\alpha)[D(\pi_o,\pi_{\beta})+ D(\pi_o,\pi_c)- D(\pi_c,\pi_{\beta})]\\
    >& 0,
\end{align*}
and $\Delta_{\beta}$ increases with $\lambda$, 
which implies that
\begin{align*}
    J(\mathcal{M},\pi_o)\geq J(\mathcal{M},\pi_{\beta})+\xi_2=J(\mathcal{M},\pi_{\beta})-\eta_1^{c}-\eta_3^{\beta}+\Delta_{\beta}>J(\mathcal{M},\pi_{\beta})
\end{align*}
for an appropriate choice of $\lambda$.

\end{document}